\documentclass{article}

\usepackage{microtype}
\usepackage{graphicx}
\usepackage{subfigure}
\usepackage{booktabs} 

\usepackage{hyperref}

\usepackage[accepted]{icml2025}

\usepackage{amsmath}
\usepackage{amssymb}
\usepackage{mathtools}
\usepackage{amsthm}

\usepackage[capitalize,noabbrev]{cleveref}

\theoremstyle{plain}

\theoremstyle{definition}

\theoremstyle{remark}

\usepackage{enumitem}
\usepackage[most]{tcolorbox}
\usepackage{verbatim}
\usepackage{amssymb}
\usepackage{bbm}
\usepackage{hyperref}
\usepackage{xspace}
\usepackage{fontawesome}
\usepackage{multirow}
\usepackage{array}
\newcommand{\PreserveBackslash}[1]{\let\temp=\\#1\let\\=\temp}
\newcolumntype{C}[1]{>{\PreserveBackslash\centering}p{#1}}
\newcolumntype{R}[1]{>{\PreserveBackslash\raggedleft}p{#1}}
\newcolumntype{L}[1]{>{\PreserveBackslash\raggedright}p{#1}}
\usepackage[textsize=tiny]{todonotes}

\usepackage{booktabs} 
\usepackage{algorithmic}
\usepackage[utf8]{inputenc} 
\usepackage[T1]{fontenc}    
\usepackage{hyperref}       
\usepackage{url}            
\usepackage{booktabs}       
\usepackage{amsfonts}       
\usepackage{nicefrac}       
\usepackage{microtype}      

\usepackage[utf8]{inputenc} 
\usepackage[T1]{fontenc}    
\usepackage{hyperref}       
\usepackage{url}            
\usepackage{booktabs}       
\usepackage{amsfonts}       
\usepackage{nicefrac}       
\usepackage{microtype}      
\usepackage{xcolor}         
\usepackage{times}
\usepackage{latexsym}
\usepackage{amsmath}
\usepackage{bm}
\usepackage{bbm}
\usepackage{multicol}
\usepackage{graphicx}
\usepackage{subfigure}
\usepackage{multirow}
\usepackage{wrapfig}
\usepackage{xcolor}

\usepackage{listings}

\definecolor{myPink}{RGB}{255,20,147} 

\newcommand{\iconlink}[3]{
  \href{#2}{\raisebox{-0.2ex}{#1}\:\textcolor{myPink}{#3}}\xspace
}

\usepackage{graphicx} 

\newcommand{\huggingface}[3]{%
  \href{#1}{\raisebox{-0.6ex}{\includegraphics[height=2.5ex]{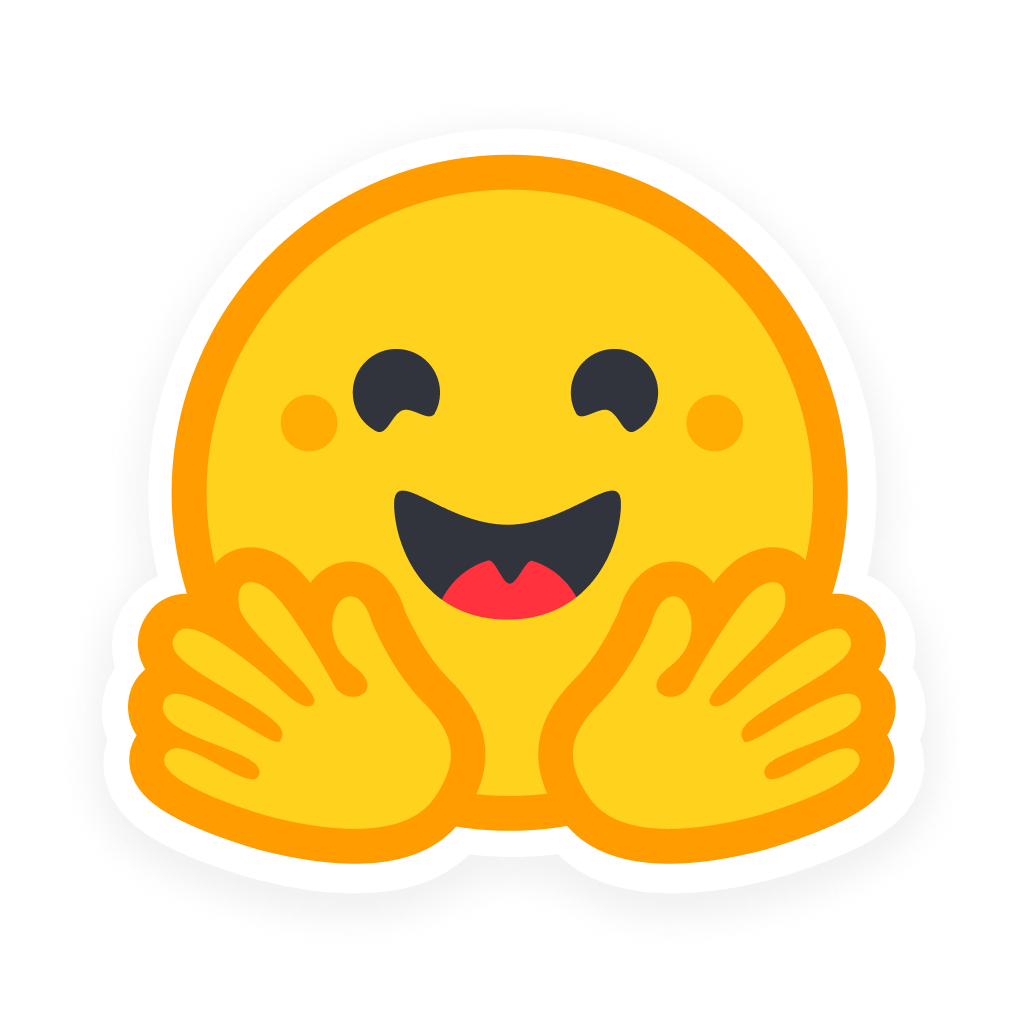}}\:\textcolor{#2}{#3}}\xspace
}

\newcommand{\dataset}[3]{%
  \href{#1}{\raisebox{-0.6ex}{\includegraphics[height=2.3ex]{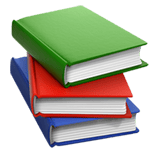}}\:\textcolor{#2}{#3}}\xspace
}

\newcommand{\homepage}[3]{%
  \href{#1}{\raisebox{-0.6ex}{\includegraphics[height=2.3ex]{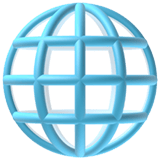}}\:\textcolor{#2}{#3}}\xspace
}

\usetikzlibrary{tikzmark}
\makeatletter
\newcommand*\myfontsize{%
  \@setfontsize\myfontsize{7}{8}%
}
\makeatother
\newcommand{\mytextbox}[2]{\tikzmarknode[draw=#1,thick,inner sep=2pt]{test}{\myfontsize #2}}
\usepackage{xcolor}
\definecolor{myred}{RGB}{220,50,50}
\definecolor{myblue}{RGB}{50,100,220}
\definecolor{mygreen}{RGB}{50,180,50}
\definecolor{mybrown}{RGB}{150,75,0}
\definecolor{mypurple}{RGB}{120,50,200}

\newcommand{\red}[1]{\mytextbox{myred}{\textbf{\textcolor{myred}{#1}}}}
\newcommand{\blue}[1]{\mytextbox{myblue}{\textbf{\textcolor{myblue}{#1}}}}
\newcommand{\green}[1]{\mytextbox{mygreen}{\textbf{\textcolor{mygreen}{#1}}}}
\newcommand{\brown}[1]{\mytextbox{mybrown}{\textbf{\textcolor{mybrown}{#1}}}}
\newcommand{\purple}[1]{\mytextbox{mypurple}{\textbf{\textcolor{mypurple}{#1}}}}
\icmltitlerunning{DeepAnalyze: Agentic Large Language Models for Autonomous Data Science}

\begin{document}

\twocolumn[
\icmltitle{DeepAnalyze: Agentic Large Language Models for Autonomous Data Science}




\begin{icmlauthorlist}
\icmlauthor{Shaolei Zhang}{ruc}
\icmlauthor{Ju Fan}{ruc}
\icmlauthor{Meihao Fan}{ruc}
\icmlauthor{Guoliang Li}{tsinghua}
\icmlauthor{Xiaoyong Du}{ruc}


\end{icmlauthorlist}

\begin{center}
\vspace{2mm}
    \iconlink{\faGithub}{https://github.com/ruc-datalab/DeepAnalyze}{ruc-datalab/DeepAnalyze}$\;$
    \huggingface{https://huggingface.co/RUC-DataLab/DeepAnalyze-8B}{myPink}{DeepAnalyze-8B}$\;$
    \dataset{https://huggingface.co/datasets/RUC-DataLab/DataScience-Instruct-500K}{myPink}{DataScience-Instruct-500K}$\;$
    \homepage{https://ruc-deepanalyze.github.io/}{myPink}{ruc-deepanalyze.github.io}
\vspace{-2mm}

\end{center}

\icmlaffiliation{ruc}{Renmin University of China}
\icmlaffiliation{tsinghua}{Tsinghua University}

\icmlcorrespondingauthor{Shaolei Zhang}{zhangshaolei98@ruc.edu.cn}
\icmlcorrespondingauthor{Ju Fan (corresponding author)}{fanj@ruc.edu.cn}


\icmlkeywords{Machine Learning, ICML}

\vskip 0.3in
]



\printAffiliationsAndNotice{} 

\newcommand{\stitle}[1]{\noindent{\bf #1}}
\newcommand{\etitle}[1]{\noindent{\underline{\em #1}}}
\newcommand{\ie}{{\em i.e.,}\xspace}
\newcommand{\eg}{{\em e.g.,}\xspace}
\newcommand{\wrt}{\emph{w.r.t.}\xspace}
\newcommand{\aka}{\emph{a.k.a.}\xspace}
\newcommand{\kwlog}{\emph{w.l.o.g.}\xspace}

\newcommand{\fanj}[1]{{\color{red} @fanj: #1}}
\newcommand{\oursys}{$\texttt{DeepAnalyze}$\xspace}
\begin{abstract}
Autonomous data science, from raw data sources to analyst-grade deep research reports, has been a long-standing challenge, and is now becoming feasible with the emergence of powerful large language models (LLMs).
Recent workflow-based data agents have shown promising results on specific data tasks but remain fundamentally limited in achieving fully autonomous data science due to their reliance on predefined workflows.
%
%
In this paper, we introduce \emph{DeepAnalyze-8B}, the first agentic LLM designed for autonomous data science, capable of automatically completing the end-to-end pipeline from data sources to analyst-grade deep research reports.
%
%
To tackle high-complexity data science tasks, we propose a curriculum-based agentic training paradigm that emulates the learning trajectory of human data scientists, enabling LLMs to progressively acquire and integrate multiple capabilities in real-world environments. We also introduce a data-grounded trajectory synthesis framework that constructs high-quality training data. Through agentic training, \textit{DeepAnalyze} learns to perform a broad spectrum of data tasks, ranging from data question answering and specialized analytical tasks to open-ended data research.
%
%
Experiments demonstrate that, with only 8B parameters, DeepAnalyze outperforms previous workflow-based agents built on most advanced proprietary LLMs. The model, code, and training data of DeepAnalyze are open-sourced, paving the way toward autonomous data science.
\end{abstract}
\vspace{-4mm}
\section{Introduction}
\label{submission}

Autonomous data science~\citep{de2022automating,sun2025surveylargelanguagemodelbased,wang2025largelanguagemodelbaseddata}, a long-standing central goal of the data science community, aims to automate the \emph{entire data science pipeline} for extracting insights from structured data.
This pipeline is inherently complex, consisting of a series of interdependent data-centric tasks spanning data preparation, analysis, modeling, visualization, and report generation.
The emergence of open-ended data research further elevates the level of complexity, going far beyond traditional question answering or task-specific analytics.
%
%
Fortunately, recent advances in large language models (LLMs) have demonstrated impressive problem-solving abilities~\citep{gpt4,gpt-o1,deepseekr1}, reshaping paradigms in domains such as search~\citep{zheng2025deepresearcherscalingdeepresearch,search-r1} and mathematics~\citep{zhang2024llamaberrypairwiseoptimizationo1like,ren2025deepseekproverv2advancingformalmathematical}.
However, despite their success on unstructured data (e.g., textual queries or contexts), LLMs still struggle to orchestrate complex, multi-stage data science pipelines and handle diverse structured data, making it difficult to achieve a general solution that works across all data science tasks.


\begin{figure}[t]
\centering
\includegraphics[width=\columnwidth]{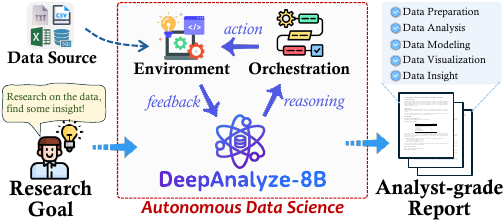}
\vspace{-6mm}
\caption{DeepAnalyze-8B is the first end-to-end agentic LLM that achieves autonomous data science, supporting entire data science pipeline and open-ended data research.}
\label{fig:ill}
\vspace{-4mm}
\end{figure}

Addressing these challenges requires endowing LLMs with two higher-level capabilities: autonomous orchestration and adaptive optimization.
First, autonomous orchestration enables LLMs to comprehend user intents and systematically coordinate a sequence of interdependent actions to accomplish complex tasks~\cite{Sapkota_2026}.
Second, adaptive optimization allows LLMs to interact with real-world data environments and iteratively refine their actions based on feedback~\cite{hong2024datainterpreterllmagent}.
As shown in Figure \ref{fig:ill}, equipped with these two capabilities, an intelligent system can robustly handle a broad spectrum of data tasks, ranging from conventional question answering and task-specific analytics to fully autonomous, open-ended data research.

%

\begin{figure*}[t]
\begin{center}
\centerline{\includegraphics[width=0.985\linewidth]{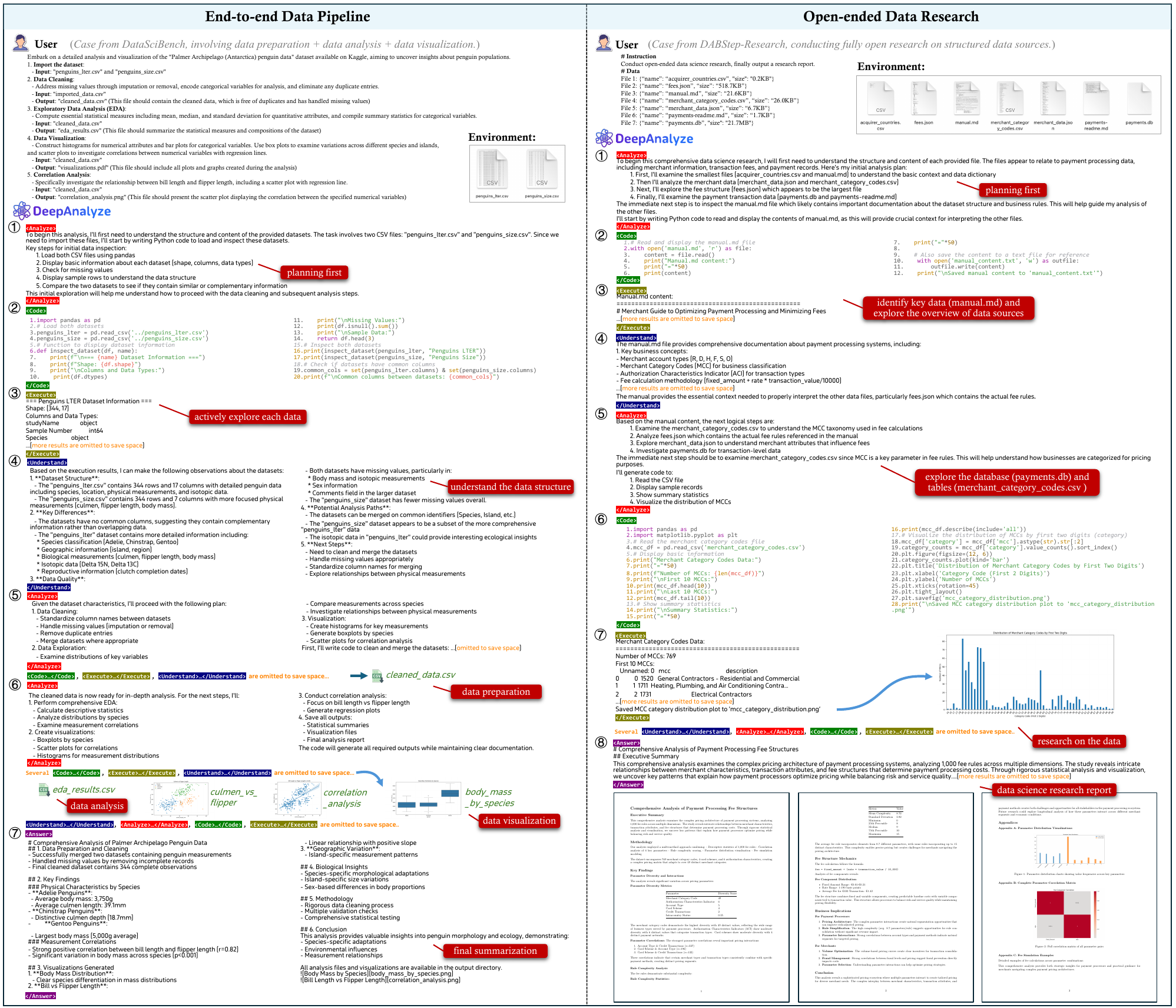}}
\vspace{-2mm}
\caption{Examples of DeepAnalyze-8B. Given the instructions and data sources in the environment, DeepAnalyze can autonomously orchestrate and optimize actions to complete a data science pipeline (left) and open-ended data research (right). DeepAnalyze first performs planning, then interacts with the data in the environment, and subsequently optimizes its actions based on feedback, ultimately accomplishing the data-centric tasks. Many intermediate actions are omitted to save space.}
\label{fig:case}
\end{center}
\vspace{-8mm}
\end{figure*}

Existing approaches to applying LLMs for autonomous data science can be broadly categorized into \emph{domain-specific LLMs} and \emph{workflow-based agents}.
Early efforts focus on developing domain-specific LLMs, such as code-oriented models~\citep{nascimento2024llm4dsevaluatinglargelanguage,wen2024groundingdatasciencecode} and structured data-oriented models~\citep{li2023tablegpttabletunedgptdiverse,jiang-etal-2023-structgpt,xu2025llasalargelanguagestructured}, to handle individual tasks like question answering or specific analytical operations.
However, these models lack the capabilities for autonomous orchestration and adaptive optimization~\citep{yang2021autopipelinesynthesizingcomplexdata,li2023autotablessynthesizingmultisteptransformations}, limiting their ability to execute the entire data science pipeline.
%
%
More recently, a line of work has explored workflow-based data science agents~\citep{NEURIPS2023_8c2df4c3,dsagent,sun2024lambda,hong-etal-2025-data}, which rely on predefined procedural workflows to prompt closed-source LLMs (e.g., GPT-4~\citep{gpt4}) to complete complex tasks.
Although these systems demonstrate stronger task coordination, they depend heavily on manually designed heuristics and domain-specific rules, falling short of achieving autonomous and adaptive behavior.
%

In essence, both domain-specific models and workflow-based agents remain limited, as they are not trained in interactive, real-world environments.
Consequently, they struggle to perform complex tasks through autonomous orchestration and adaptive optimization.
Notably, recent advances in agentic training, a new training paradigm successfully applied in the search domain~\citep{zheng2025deepresearcherscalingdeepresearch,search-r1}, have demonstrated that reinforcement learning in real-world environments is crucial for enabling LLMs to develop autonomous problem-solving capabilities.

In this paper, we aim to advance LLM-based data science methods from workflow-based agents to a trainable agentic model that learns to autonomously perform data science tasks in real-world environments.
However, applying agentic training to this domain presents two key challenges: \emph{reward sparsity} and \emph{trajectory scarcity}.
On the one hand, the inherent complexity of data science tasks makes it difficult for foundation LLMs to complete tasks successfully during the early stages of training. This leads to severe reward sparsity, i.e., a lack of positive reinforcement signals, which can hinder or even collapse the entire agentic training process.
On the other hand, the scarcity of long-chain problem-solving trajectories in data science provides insufficient guidance for LLMs to explore the solution space effectively, resulting in inefficient, blind trial-and-error exploration without meaningful intermediate supervision.
%

To address these challenges, we introduce DeepAnalyze, an agentic LLM designed for autonomous data science.
As illustrated in Figure~\ref{fig:case}, with only 8B parameters, DeepAnalyze can automate the entire data science pipeline, ranging from specific data tasks to open-ended data research, providing a unified and general solution for data-centric applications.
%
Specifically, to mitigate reward sparsity, DeepAnalyze adopts a curriculum-based agentic training paradigm inspired by the learning trajectory of human data scientists. This progressive easy-to-difficult schedule enables the model to gradually evolve from mastering individual skills to developing comprehensive, adaptive problem-solving abilities in real-world environments.
To address trajectory scarcity, we propose a data-grounded trajectory synthesis framework that automatically constructs high-quality reasoning and interaction trajectories, offering effective exploration guidance within the large solution space.
Through this combination of curriculum-based training and trajectory synthesis, DeepAnalyze learns to autonomously orchestrate actions and adaptively optimize its strategies, enabling it to tackle complex and diverse data science tasks effectively.

%

In summary, our key contributions are three-fold.
\begin{itemize}[leftmargin=1em, itemsep=-1ex, topsep=0ex]
	\item \textbf{Agentic Model:} To the best of our knowledge, DeepAnalyze is the first agentic LLM tailored for autonomous data science, endowed with two indispensable capabilities, \emph{autonomous orchestration} and \emph{adaptive optimization}. DeepAnalyze serves as a \emph{foundation model} that can be directly applied or further customized through prompting or supervised fine-tuning for specific scenarios.
    %
    
	\item \textbf{Agentic Training}: We propose a curriculum-based agentic training paradigm with data-grounded trajectory synthesis to address \emph{reward sparsity} and \emph{trajectory scarcity}, enabling effective learning for high-complexity tasks that require multiple abilities.
	\item \textbf{Strong Performance}: 
    Experimental results on 12 benchmarks show that, with only 8B parameters, \textbf{DeepAnalyze-8B} surpasses most advanced proprietary LLMs. More importantly, it is the first agentic model capable of performing open-ended data research and generating analyst-grade reports.
\end{itemize}

\section{Related Work}

\textbf{Autonomous Data Science.}\quad 
Autonomous data science has long been pursued as an important goal of intelligent systems. Existing LLM-based data science methods can be categorized into: domain-specific LLMs and workflow-based agents. To handle individual tasks in data science, early methods focused on fine-tuning LLMs into domain-specific models, including LLMs for data science code generation \citep{nascimento2024llm4dsevaluatinglargelanguage,wen2024groundingdatasciencecode,nejjar2024llmsscienceusagecode,pan2025visshepherdconstructingcriticllmbased}, tabular LLMs \citep{li2023tablegpttabletunedgptdiverse,fang2024large,zhang-etal-2025-tablellm,xu2025llasalargelanguagestructured,nvAgent2025ACL,lei2025reasoningtableexploringreinforcementlearning}, and database-oriented LLMs \citep{xue2024dbgptempoweringdatabaseinteractions,liu2024survey,mohammadjafari2025naturallanguagesqlreview}. Recently, an increasing number of data agents have demonstrated promising performance in data science by leveraging workflows to gradually prompt LLMs for complex tasks \citep{NEURIPS2023_8c2df4c3,dsagent,MatPlotAgent2024ACL,sun2024lambda,hong-etal-2025-data}. Most existing agents are built upon Chain-of-Thought frameworks, including ReAct \citep{yao2023react}, AutoGen \citep{wu2024autogen}, and self-reflection \citep{pan2023automaticallycorrectinglargelanguage}, which decompose complex tasks into multiple subtasks and solve them sequentially. Regardless of workflow design, existing agents primarily rely on carefully crafted prompting to guide closed-source LLMs in performing data science tasks.


\begin{figure*}[t]
	\begin{center}
		\centerline{\includegraphics[width=\linewidth]{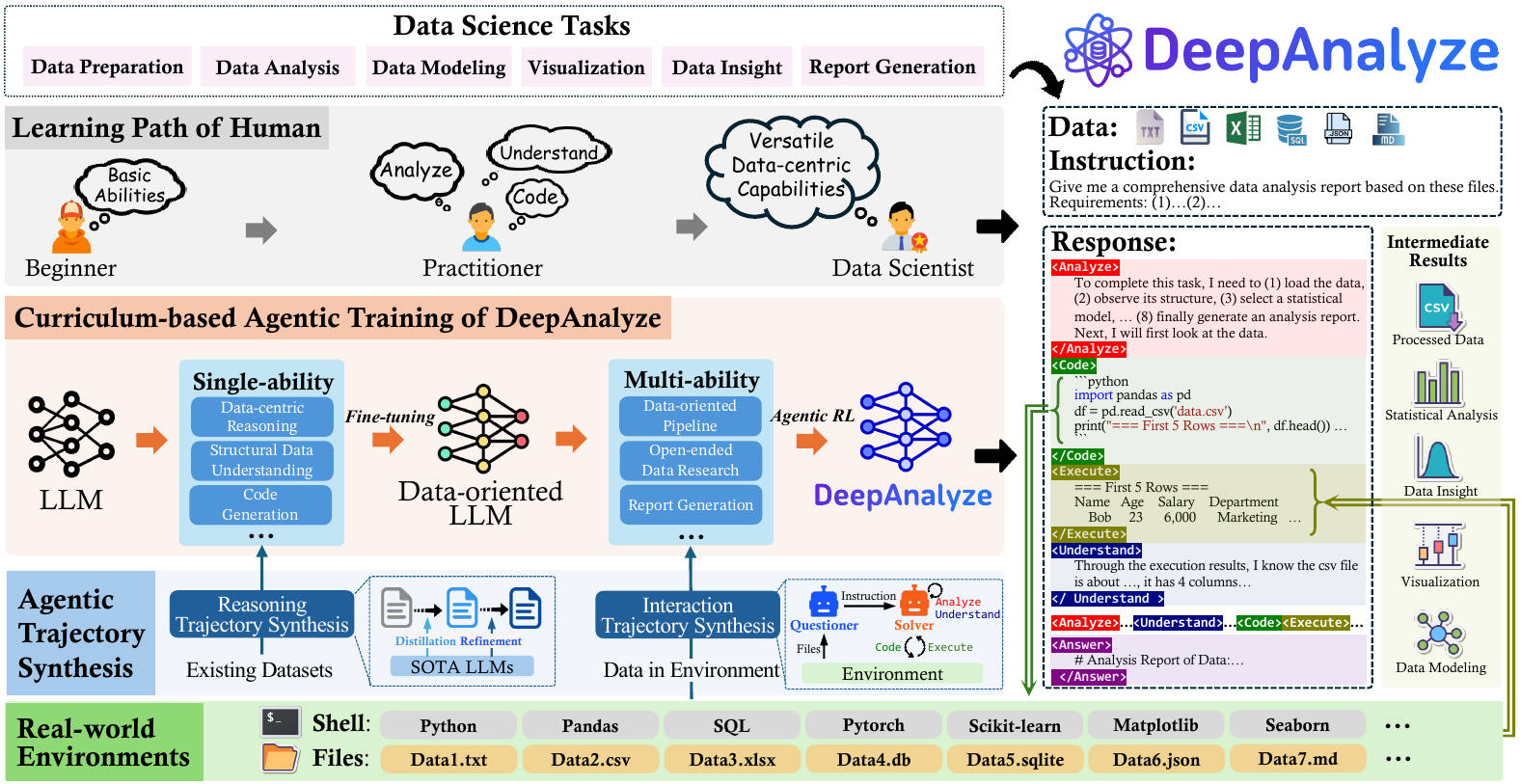}}
		\vspace{-2mm}
		\caption{Architecture of DeepAnalyze.}
		\label{fig:model}
	\end{center}
	\vspace{-8mm}
\end{figure*}

Despite these advances, domain-specific LLMs (focused on individual tasks) and workflow-based agents (dependent on manually designed workflows) remain incapable of fully autonomous data science. Therefore, the proposed DeepAnalyze does not rely on prompting frameworks or predefined workflows, instead, it internalizes data science capabilities through agentic training within real-world environments.


\textbf{Agentic Training for LLM.}\quad Agentic training aims to enhance LLMs as agentic models through reinforcement learning and thereby enable LLMs to perform multi-step reasoning and interactions in real-world environments \citep{plaat2025agenticlargelanguagemodels}, which has already achieved practical success in coding \citep{sapkota2025vibecodingvsagentic} and searching \citep{zheng2025deepresearcherscalingdeepresearch,li2025searcho1agenticsearchenhancedlarge,search-r1}. Typically, these methods use prompts to control the interaction format of LLMs and complete RL with the accuracy of the final answer as the reward. Based on this, lightweight cold-start is proposed to help LLMs learn the interaction format \citep{deepseekr1}, improving the initial state for RL training. Existing training methods mainly focus on reasoning ability, while data science requires a broader range of abilities, such as reasoning, structured data understanding, and code generation. This complexity makes that initial LLMs (even after cold-start format learning) are generally incapable of completing complex data science tasks, leading to challanges of reward sparsity and trajectory scarcity. To this end, we propose a curriculum-based agentic training that enables LLMs to gradually acquire complex data science skills through a progression from single to multiple abilities, while employing data-grounded trajectory synthesis to generate high-quality reasoning and interaction trajectories for training.


\section{DeepAnalyze}


In this paper, we introduce DeepAnalyze, an agentic large language model for autonomous data science. To endow the LLM with the capability for autonomous orchestration and adaptive optimization in real-world environments, we propose a curriculum-based agentic training and data-grounded trajectory synthesis framework tailored for complex tasks with multiple abilities. Specifically, inspired by the behavior of human data scientists, we first define a set of actions that enable DeepAnalyze to directly interact with the data in its environment. Building on this architecture, we automatically synthesize high-quality data science trajectories and introduce a curriculum-based agentic training paradigm that guides DeepAnalyze through a progression from a beginner to data scientist, thereby empowering DeepAnalyze to tackle a wide spectrum of data science tasks. The architecture, curriculum-based agentic training, and data-grounded trajectory synthesis are introduced as follows.

\subsection{Architecture}

Unlike foundation LLMs that focus on understanding and generating natural language, LLMs for data science meet the additional challenge of understanding and interaction with structured data, which is typically stored in external files. Therefore, DeepAnalyze extends natural language interaction by introducing data-oriented interaction pattern, thereby enabling LLMs to autonomously interact with real-world environments.

\textbf{Inputs Format.}\quad Previous structured data–specific LLMs \citep{li2023tablegpttabletunedgptdiverse,fang2024large,zhang-etal-2025-tablellm,xu2025llasalargelanguagestructured,lei2025reasoningtableexploringreinforcementlearning} often converted tables stored in databases, CSV, or XLSX files into unstructured Markdown text and fed them into the LLM's context to enable structured data understanding. However, due to context length limitations, these methods can only handle small-scale data (e.g., very small tables).
When human data scientists work with large-scale data, they do not passively read and memorize every record. Instead, they actively explore each data source as needed and then plan the following steps accordingly. To this end, DeepAnalyze integrates both modes: it passively accepts structured data expressed as text in the input, while also actively inspecting external data sources according to user inputs, where the filenames of the external data sources are specified in inputs, as shown in Figure \ref{fig:case}.

		

\textbf{Interaction Pattern.}\quad
Given an instruction and the data sources in the environment, data scientists typically analyze, interact with the data in the environment, understand structured data, and iterate until the instruction is completed. To emulate this process, DeepAnalyze introduces five actions to automatically accomplish the data science task, including:
\begin{itemize}[leftmargin=1em, itemsep=-1ex, topsep=0ex]
	\item  \red{$\langle$Analyze$\rangle$}$\cdots$\red{$\langle$/Analyze$\rangle$}: Analyze textually, including planning, reasoning, reflection, self-verification...
	\item  \blue{$\langle$Understand$\rangle$}$\cdots$\blue{$\langle$/Understand$\rangle$}: Understand the content of data source, such as databases, tables, and documents.
	\item  \green{$\langle$Code$\rangle$}$\cdots$\green{$\langle$/Code$\rangle$}: Generate code to interact with the data in the environment, using Python suited for data science.
	\item  \brown{$\langle$Execute$\rangle$}$\cdots$\brown{$\langle$/Execute$\rangle$}: Execute code and collect the feedback from the environment.
	\item  \purple{$\langle$Answer$\rangle$}$\cdots$\purple{$\langle$/Answer$\rangle$}: Produce the final output.
\end{itemize}
In practice, we extend the vocabulary of the foundation LLM to support the generation of these special tokens. During the inference, DeepAnalyze automatically switches between different actions by generating these special tokens, as shown in the right side of Figure \ref{fig:model}. In particular, once a \green{$\langle$Code$\rangle$}$\cdots$\green{$\langle$/Code$\rangle$} is generated, DeepAnalyze executes the code in the environment and places the feedback in \brown{$\langle$Execute$\rangle$}$\cdots$\brown{$\langle$/Execute$\rangle$}, and then generates the next action. The detailed inference process of DeepAnalyze is shown in Algorithm \ref{alg:DeepAnalyze}. With this architecture, all actions (i.e., special tokens) are autonomously generated by the model without any human-defined workflows or rules, which allows DeepAnalyze to fully autonomously orchestrate and optimize each action, laying the foundation for autonomous data science.

\begin{algorithm}[t]
\caption{Inference of DeepAnalyze}\label{alg:DeepAnalyze}
\footnotesize
\begin{algorithmic}[1]
   \STATE {\bfseries Input:} Instruction $Q$, Environment $Env$, DeepAnalyze model $\mathcal{M}$
   \STATE {\bfseries Output:} Response $A$ (with interaction process)
   \STATE {\bfseries Initialization:} $A = \emptyset$
   \WHILE{\purple{$\langle$Answer$\rangle$}$\cdots$\purple{$\langle$/Answer$\rangle$}  not in $A$}
       \STATE $y \gets \mathcal{M}(Q, A)$ $\;\;\;\;\;\;$\textcolor{gray}{// generate next action based on the instruction $Q$ and current response $A$}
       \STATE $A \gets A + y$
       \IF{\green{$\langle$Code$\rangle$}$\cdots$\green{$\langle$/Code$\rangle$} in $y$}
           \STATE $code$ $\gets$ extract\_code$(y)$
           \STATE $feedback$ $\gets Env$.execute(code) $\;\;\;$\textcolor{gray}{// interaction with the data in the environment}
           \STATE $A \gets A + $\brown{$\langle$Execute$\rangle$} $+feedback+$ \brown{$\langle$/Execute$\rangle$}
       \ENDIF
   \ENDWHILE
   \STATE \textbf{Return} $A$
\end{algorithmic}
\end{algorithm}

\subsection{Curriculum-based Agentic Training}


Under the above architecture, DeepAnalyze need to learn how to interact with the environment to accomplish various data science tasks. Unlike individual coding or searching task, data science tasks demand a broader and more complex set of abilities, ranging from reasoning, structured data understanding, and code generation to the composite abilities needed for entire data science pipeline and open-ended research. The complexity of these capabilities results in the limited proficiency of foundation LLMs in data science domains \citep{zhang2025datascibenchllmagentbenchmark}, leading to severe reward sparsity on complex tasks and rendering existing agentic training (such as RL-Zero or RL with cold-start training \citep{deepseekr1}) ineffective due to the lack of positive feedback. To address this challenge, we propose curriculum-based agentic training, which emulates the learning path of human data scientists by gradually transitioning from mastering single abilities to integrating multiple abilities. This training framework consists of two stages, where stage 1 employs single-ability fine-tuning to strengthen the foundation LLM's single ability, and stage 2 uses multi-ability agentic training to enable the LLM to apply multiple abilities in real-world environments to accomplish complex data science tasks.

\begin{figure}[t]
	\begin{center}
		\centerline{\includegraphics[width=\columnwidth]{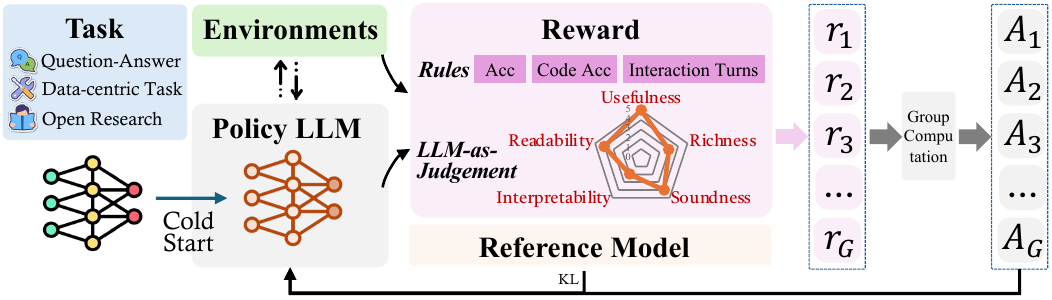}}
		\vspace{-2mm}
		\caption{Schematic diagram of agentic RL.}
		\label{fig:ill}
	\end{center}
	\vspace{-8mm}
\end{figure}

\begin{figure*}[t]
	\centering
	\subfigure[Reasoning Trajectory Synthesis]{
		\includegraphics[width=0.47\linewidth]{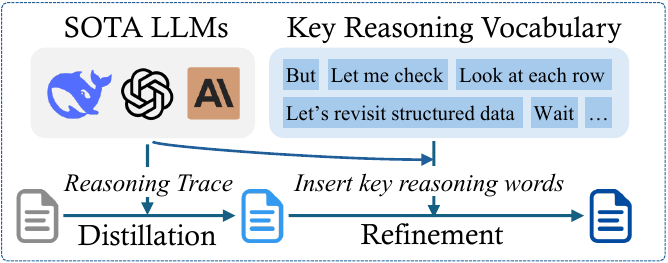} \label{fig:data1}
	}
	\subfigure[Interaction Trajectory Synthesis]{
		\includegraphics[width=0.47\linewidth]{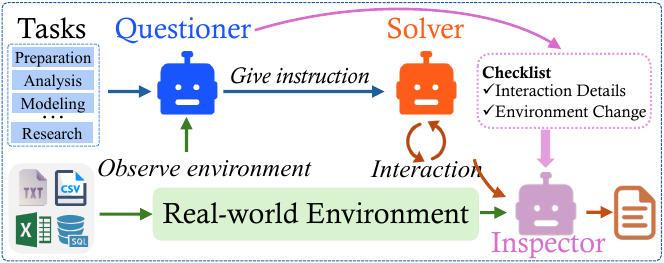} \label{fig:data2}
	}
	\vspace{-2mm}
	\caption{The proposed data-grounded trajectory synthesis for the development of DeepAnalyze on data science tasks.}
	\label{fig:data}
	\vspace{-1mm}
\end{figure*}

\textbf{Single-ability Fine-tuning.}\quad
Since most foundation LLMs have not been trained specifically for data science tasks, in this stage, we first enhance the various single abilities that data science relies, primarily including reasoning, structured data understanding, and code generation, which correspond respectively to the actions $\langle$Analyze$\rangle$, $\langle$Understand$\rangle$, and $\langle$Code$\rangle$. Specifically, we fine-tune the foundation LLM using long CoT data (i.e., including reasoning traces) of general tasks, structured data understanding, code generation. This stage of training mirrors the human learning process from a beginner to a data science practitioner in acquiring specialized skills, enhancing LLM's single ability in various aspects of data science.

\textbf{Multi-ability Agentic Training.}\quad
Building on the mastery of various single abilities, we employ agentic reinforcement learning to train DeepAnalyze to apply multiple abilities in real-world environments to complete complex data science tasks. To ensure the quality of reinforcement learning, we first perform a cold start by fine-tuning the LLM on synthesized interaction trajectories, enabling it to acquire basic capabilities in orchestrating and optimizing individual actions. Subsequently, we train DeepAnalyze in real-world environments using reinforcement learning with group relative policy optimization (GRPO) \citep{shao2024deepseekmathpushinglimitsmathematical}. For each question $q$ in training data $D$, GRPO samples a group of $G$ outputs $\{o_1,\cdots,o_G\}$ from the old policy $\pi_{\theta_{\mathrm{old}}}$ and then optimizes the policy model $\pi_\theta$ by maximizing the following objective:
\begin{equation}
	\begin{split}
		\mathcal{J}_{\mathrm{GRPO}}(\theta) =
		\mathbb{E}_{q\thicksim D,\{o_i\}_{i=1}^G \thicksim \pi_{\theta_{\mathrm{old}}}(\cdot|q)}
		\Biggl[ 
		\frac{1}{G}\sum_{i=1}^G \Bigl(
		\min\Bigl(\\
		\frac{\pi_\theta(o_i|q)}{\pi_{\theta_{\mathrm{old}}}(o_i|q)} A_i, 
		\mathrm{clip}\Bigl(\frac{\pi_\theta(o_i|q)}{\pi_{\theta_{\mathrm{old}}}(o_i|q)}, 1-\varepsilon, 1+\varepsilon\Bigr) A_i
		\Bigr) \\
		- \beta \mathrm{D}_{KL}\left(\pi_\theta \parallel \pi_{\mathrm{ref}}\right)
		\Bigr)
		\Biggr]
	\end{split}
\end{equation}
where $A_i$ is the advantage calculated from the rewards $\{r_1,\cdots,r_G\}$ of outputs within each group, $\pi_{\mathrm{ref}}$ is the reference model, $\varepsilon$ and $\beta$ are hyperparameters.

\textbf{Hybrid Reward Modeling.}\quad 
The effectiveness of agentic reinforcement learning critically depends on both the training data and the reward function. We use the agentic interaction trajectories synthesized in Section \ref{sec:Agentic Data Synthesis} as training data, covering three broad categories of data science tasks: data question answering, specific data tasks (e.g., data preparation, analysis, visualization, modeling, and insight extraction), and open-ended research. Since many data science tasks are inherently open-ended, we adopt a hybrid reward modeling that combines rule-based rewards with LLM-as-a-judge rewards. For all tasks, we first check whether the output format conforms to DeepAnalyze's architecture (i.e., whether it contains exactly five types of actions with the correct format). If the format is incorrect, we directly assign a reward of $R=-1$.

For data question answering and data-centric tasks, which have reference answers, the reward $R$ of each output $o$ are calculated using accuracy and interaction trajectory quality:
\begin{gather}
	R=\frac{1}{2}(\mathbbm{1}_{acc}(o)+{S}_{interaction}(o))
\end{gather}
where $\mathbbm{1}_{acc}(o)\in\{0,1\}$ indicates whether the result is correct, and ${S}_{interaction}(o)\in [0,1]$ is a score to evaluate the quality of the interaction trajectory.

For open-ended research, the reward $R$ of each output $o$ is evaluated based on the quality of the final research report and the research process. Denoting each interaction turn in output $o$ as $T_i\in o$, the reward $R$ is calculate as:
\begin{gather}
	R=\frac{1}{3}\!\left(\!{S}_{report}(o)\!+\!\min(\frac{|T|}{N^{T}},1)\!+\!\frac{1}{|T|}\sum_{T_i\in o}\mathbbm{1}_{success}(T_{i}) \!\right)
\end{gather}
where ${S}_{report}(o)$ is the score that evaluates the generated report from five aspects: usefulness, richness, soundness, interpretability, and readability. $|T|$ measures the interaction turns with the environment, where $N^{T}=10$ is a hyperparameter. $\mathbbm{1}_{success}(T_{i})$ indicates whether each interaction turn is successful. This reward encourages DeepAnalyze to engage in more successful interactions with the environment and to generate high-quality research report.

Through curriculum-based agentic training, we progressively enhances DeepAnalyze's capabilities following an easy-to-hard schedule, ultimately enabling it to autonomously accomplish a variety of data science tasks in real-world environments.

\subsection{Data-grounded Trajectory Synthesis}
\label{sec:Agentic Data Synthesis}

The proposed curriculum-based agentic training relies on high-quality reasoning and interaction trajectory data, while such data is unfortunately scarce for data science tasks. To overcome this challange, we introduce a data-grounded trajectory synthesis framework that automatically constructs high-quality trajectory data tailored for data science tasks. The data-grounded trajectory synthesis framework consists of two parts: \emph{Reasoning Trajectory Synthesis}, which construct the reasoning trajectory for existing structured data instruction datasets, and \emph{Interaction Trajectory Synthesis}, which constructs entire data science trajectory based on structured data sources in the environment.

\begin{table*}[t]
	\scriptsize
	\centering
	\begin{tabular}{lccccccccc}\toprule
		\multirow{4}{*}{\textbf{Models}}      & \multicolumn{2}{c}{\textbf{Coarse-grained Metrics}}                                                                                  & \multicolumn{6}{c}{\textbf{Fine-grained Metrics}}                                                                                                                                                                                                                                                                                                                                            & \multirow{4}{*}{\textbf{Score}} \\ \cmidrule(lr){2-3}\cmidrule(lr){4-9}
		& \textbf{\begin{tabular}[c]{@{}c@{}}Success\\ Rate\end{tabular}} & \textbf{\begin{tabular}[c]{@{}c@{}}Completion\\ Rate\end{tabular}} & \textbf{VLM}  & \textbf{\begin{tabular}[c]{@{}c@{}}F1: Data \\ Preparation\end{tabular}} & \textbf{\begin{tabular}[c]{@{}c@{}}F2: Plot\\ Validity\end{tabular}} & \textbf{\begin{tabular}[c]{@{}c@{}}F3: Data\\ Exploration\end{tabular}} & \textbf{\begin{tabular}[c]{@{}c@{}}F4: Data\\ Visualization\end{tabular}} & \textbf{\begin{tabular}[c]{@{}c@{}}F5: Data\\ Modeling\end{tabular}} &                                 \\ \midrule
		\multicolumn{10}{c}{\textbf{\textit{Close-Source API-Based Agent}}}                                                                                                                                                                                                                                                                                                                                                                                                                                                                                                                                                             \\\midrule
		\textbf{o1-mini}                      & 29.77                                                           & 45.26                                                              & 2.87          & 44.63                                                                    & 19.27                                                                & 36.01                                                                   & 30.94                                                                     & 23.81                                                                & 38.78                           \\
		\textbf{GPT-4o-mini}                  & 50.63                                                           & 57.78                                                              & 3.05          & 60.30                                                                    & 48.02                                                                & 57.84                                                                   & 59.24                                                                     & 53.54                                                                & 54.18                           \\
		\textbf{GPT-4o}                       & \textbf{66.31}                                                  & \textbf{68.44}                                                     & \textbf{3.91} & \textbf{75.93}                                                           & \textbf{56.14}                                                       & \textbf{69.33}                                                          & \textbf{71.35}                                                            & \textbf{57.67}                                                       & \textbf{64.51}                  \\
		\textbf{GPT-4-Turbo}                  & 51.93                                                           & 58.87                                                              & 3.09          & 62.30                                                                    & 41.62                                                                & 57.75                                                                   & 60.25                                                                     & 50.75                                                                & 54.65                           \\
		\textbf{Claude-3-5-Sonnet}            & 47.48                                                           & 58.11                                                              & 2.14          & 49.07                                                                    & 36.94                                                                & 55.84                                                                   & 52.87                                                                     & 46.04                                                                & 52.29                           \\
		\textbf{GLM-4-Flash}                  & 30.32                                                           & 34.04                                                              & 1.33          & 36.53                                                                    & 29.42                                                                & 32.57                                                                   & 27.64                                                                     & 14.44                                                                & 30.74                           \\\midrule
		\multicolumn{10}{c}{\textbf{\textit{Open-Source LLM-based Agent}}}                                                                                                                                                                                                                                                                                                                                                                                                                                                                                                                                                              \\\midrule
		\textbf{Llama-3.1-8B-Instruct}        & 24.73                                                           & 33.89                                                              & 1.29          & 38.24                                                                    & 18.25                                                                & 21.98                                                                   & 22.89                                                                     & 25.85                                                                & 29.69                           \\
		\textbf{Gemma-2-9B-it}                & 7.07                                                            & 11.00                                                              & 1.06          & 26.16                                                                    & 16.90                                                                & 23.81                                                                   & 18.11                                                                     & 17.15                                                                & 12.66                           \\
		\textbf{GLM-4-9B-Chat}                & 25.72                                                           & 30.38                                                              & 1.69          & 31.51                                                                    & 23.15                                                                & 28.07                                                                   & 27.19                                                                     & 19.14                                                                & 27.57                           \\
		\textbf{Qwen2.5-7B-Instruct}          & 43.83                                                           & 50.74                                                              & 1.43          & 51.18                                                                    & 36.41                                                                & 47.25                                                                   & 45.24                                                                     & 34.77                                                                & 45.99                           \\
		\textbf{Qwen2-7B-Instruct}            & 22.84                                                           & 25.58                                                              & 1.16          & 30.93                                                                    & 20.78                                                                & 28.73                                                                   & 25.87                                                                     & 7.52                                                                 & 23.52                           \\
		\textbf{Yi-1.5-9B-Chat-16K}           & 38.20                                                           & 42.35                                                              & 0.73          & 38.14                                                                    & 36.36                                                                & 35.64                                                                   & 37.08                                                                     & 27.79                                                                & 38.22                           \\
		\textbf{CodeLlama-13B-Instruct}       & 10.49                                                           & 14.64                                                              & 0.04          & 11.67                                                                    & 11.34                                                                & 9.43                                                                    & 14.43                                                                     & 5.15                                                                 & 12.64                           \\
		\textbf{CodeLlama-7B-Instruct}        & 2.88                                                            & 3.97                                                               & 0.00          & 3.53                                                                     & 2.37                                                                 & 2.57                                                                    & 1.74                                                                      & 1.59                                                                 & 3.31                            \\
		\textbf{StarCoder2-15B}               & 2.07                                                            & 2.61                                                               & 0.07          & 2.57                                                                     & 1.81                                                                 & 1.59                                                                    & 3.43                                                                      & 1.19                                                                 & 2.33                            \\
		\textbf{Deepseek-Coder-6.7B-instruct} & 37.03                                                           & 41.62                                                              & 1.93          & 43.49                                                                    & 34.57                                                                & 46.36                                                                   & 46.49                                                                     & 18.09                                                                & 38.45                           \\
		\textbf{Qwen2.5-Coder-7B-Instruct}    & 45.18                                                           & 53.11                                                              & 1.48          & 51.58                                                                    & 43.21                                                                & 43.87                                                                   & 42.50                                                                     & \textbf{35.23}                                                       & 47.67                           \\\midrule
		\multicolumn{10}{c}{\textbf{\textit{Agentic Model}}}                                                                                                                                                                                                                                                                                                                                                                                                                                                                                                                                                                            \\\midrule
		\textbf{DeepAnalyze-8B}               & \textbf{59.91}                                                  & \textbf{66.24}                                                     & \textbf{2.86} & \textbf{71.68}                                                           & \textbf{67.86}                                                       & \textbf{58.62}                                                          & \textbf{69.09}                                                            & 33.33                                                                & \textbf{61.11}        \\\bottomrule         
	\end{tabular}
	\caption{Performance on DataSciBench. `Success Rate' and `Completion Rate' are pass rate and accuracy. `VLM' and `F1-F5' scores evaluate performance on various fine-grained data science sub-tasks, `Score' denotes the overall performance.}
	\label{tab:DataSciBench}
\end{table*}

\textbf{Reasoning Trajectory Synthesis.}\quad
Existing instruction datasets for structured data, such as TableQA \citep{li2023tablegpttabletunedgptdiverse,lei2025reasoningtableexploringreinforcementlearning}, structured knowledge grounding \citep{zhuang2024structlmbuildinggeneralistmodels}, and data science code generation, is useful to improve LLM's single capability. However, these datasets typically contain only instructions and responses, without the reasoning process. To address this limitation, we enhance existing datasets by synthesizing complex and refined reasoning trajectories, which are used for DeepAnalyze's single ability training.

As shown in Figure \ref{fig:data1}, given the instruction–response pairs in the original dataset, the reasoning trajectory synthesis involves distillation and refinement steps. In the distillation step, we employ advanced LLMs as teacher models to extract their reasoning trajectories, whose correctness is verified by comparing the generated responses with the ground truth responses \citep{deepseekr1}. To strengthen the model's understanding of structured data, the distilled reasoning is reformulated by advanced LLMs into two complementary components: $\langle$Analyze$\rangle$ (reasoning process) and $\langle$Understand$\rangle$ (structured data understanding). Building on this, we introduce keyword-guided refinement to further enhance the reasoning trajectories with a focus on structured data. Previous works have shown that certain keywords, such as ``but''/``wait'', play a crucial role in reasoning \citep{zhang2025reasoningmodelsknowtheyre,shen2025longimportantdifficulttraining}. Following this insight, we construct a key reasoning vocabulary and sample key reasoning words to insert into the reasoning trajectory, thereby improving its reasoning on structured data. Appendix \ref{app:Reasoning} provides an example of keyword-guided refinement, where inserting keywords enhances the reasoning process by focusing more on the data, thereby improving the quality of the reasoning trajectory. Through reasoning trajectory synthesis, we can effectively leverage existing datasets to improve DeepAnalyze's single ability in reasoning, structured data understanding, and code generation.

\begin{figure}[t]
	\begin{center}
		\centerline{\includegraphics[width=0.9\columnwidth]{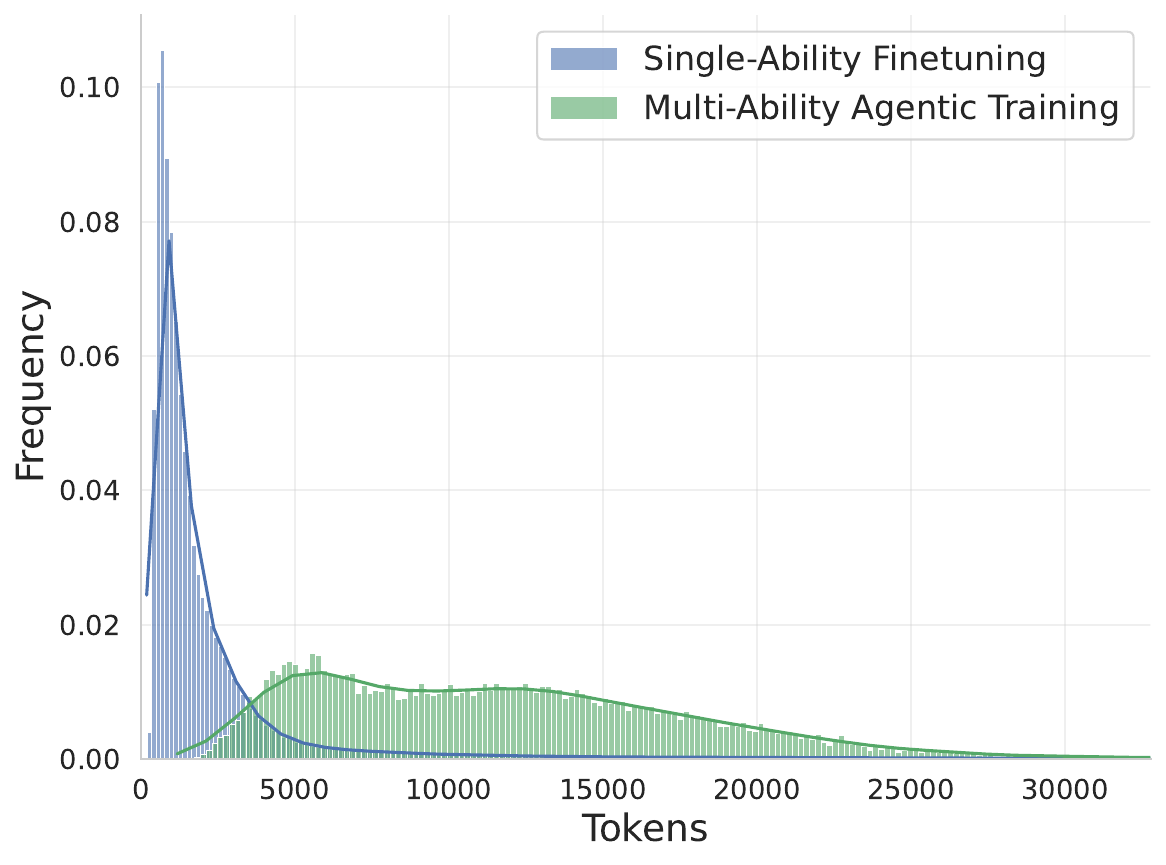}}
		\vspace{-4mm}
		\caption{Length distribution of training data.}
		\label{fig:tokens_distribution}
	\end{center}
	\vspace{-8mm}
\end{figure}

\textbf{Interaction Trajectory Synthesis.}\quad
To enable DeepAnalyze to autonomously orchestrate and optimize multiple abilities in real-world environments, it is essential to construct multi-turn interaction trajectory data with the environment, yet such data is extremely scarce. In contrast, NL2SQL datasets such as Spider \citep{yu-etal-2018-spider} and BIRD \citep{li2024can} provide abundant structured data sources. To bridge this gap, we develop a multi-agent system to synthesize data science interaction trajectories from these data sources.

The multi-agent system involves three roles: questioner, solver, and inspector. The questioner observes the data sources in the environment and accordingly formulates a data science problem, conditioned on a sampled task type (e.g., data preparation, data analysis, data modeling, data insight, or open-ended research). Simultaneously, the questioner produces a checklist that serves as the evaluation criterion, including interaction-level constraints (e.g., number of turns, code library) and environment-level constraints (e.g., whether new files are generated, detailed file name). Given the data science problem and the data sources, the solver interacts with the environment using the introduced five actions to complete the task. Finally, the inspector validates the trajectory by checking the interaction process and environmental changes against the checklist, determining whether the trajectory should be accepted. Importantly, filtering trajectories based on both interaction details and environmental changes substantially improves the quality of synthesized data. Through interaction trajectory synthesis, the high-quality multi-turn interaction data can be used for multi-ability agentic training (cold start and RL).

\subsection{DataScience-Instruct-500K}
\label{sec:DataScience-Instruct-500K}
We develop DeepAnalyze based on the constructed data in Sec.\ref{sec:Agentic Data Synthesis}. During the single-ability fine-tuning stage, we employ the reasoning trajectories built for data science, along with 100K general reasoning samples from AM-DeepSeek-R1-0528-Distilled\footnote{\url{https://huggingface.co/datasets/a-m-team/AM-DeepSeek-R1-0528-Distilled}}. In the multi-ability agentic training stage (including both cold start and RL phases), we use the interaction trajectories constructed for data science.

Figure \ref{fig:tokens_distribution} illustrates the length distribution of training data in both stages, with a sequence length of 8K in the first stage and 32K in the second. In terms of scale, the single-ability fine-tuning stage consists of approximately 470K samples, the cold-start phase of multi-ability training includes 20K samples, and the RL phase comprises 15K samples, resulting in a total of around 500K samples. We release all training data, named \texttt{DataScience-Instruct-500K}\footnote{\url{https://huggingface.co/datasets/RUC-DataLab/DataScience-Instruct-500K}}, which can be used to train LLMs for data science tasks.

\section{Experiments}

\subsection{Benchmarks}
We conduct experiments on 12 data science benchmarks.

\textbf{DataSciBench} \citep{zhang2025datascibenchllmagentbenchmark} is the latest benchmark to evaluate LLM's capabilities on the entire data science pipeline, covering data preparation, data analysis, data modeling, data visualizatoin, and data insight.

\textbf{DSBench} \citep{jing2025dsbench} evaluates data analysis and modeling capabilities, comprising 540 real-world tasks collected from ModelOff and Kaggle competitions.

\textbf{DABStep} \citep{egg2025dabstepdataagentbenchmark} is a data agent benchmark with 450 real-world data analysis tasks designed to evaluate the multi-step reasoning abilities of agents.

\textbf{DABStep-Research} is a benchmark we constructed based on DABStep \citep{egg2025dabstepdataagentbenchmark} to evaluate the capability of data science report generation. Considering that existing data science benchmarks rarely assess deep research abilities on structured data, we propose DABStep-Research to measure the capability to generate comprehensive data research reports from raw data sources. The evaluation covers five aspect: data preparation, data analysis, data insight, report generation, and open-ended data research. Please refer to Appendix \ref{app:DABStep-Research} for details on its construction and cases.

\textbf{DS-1000} \citep{pmlr-v202-lai23b} is a code generation benchmark containing 1000 data science problems spanning seven Python libraries such as NumPy, Pandas, Matplotlib, etc.

\textbf{TableQA Benchmarks} are a series of question-answering benchmarks based on structured tables, including WikiTQ \citep{pasupat-liang-2015-compositional}, HybridQA \citep{chen-etal-2020-hybridqa}, MultiHiertt \citep{zhao-etal-2022-multihiertt}, OTT-QA \citep{chen2021open}, FinQA \citep{chen-etal-2021-finqa}, TAT-QA \citep{nan-etal-2022-fetaqa}, and HiTab \citep{cheng-etal-2022-hitab}.

\begin{figure}[t]
	\begin{center}
		\centerline{\includegraphics[width=\columnwidth]{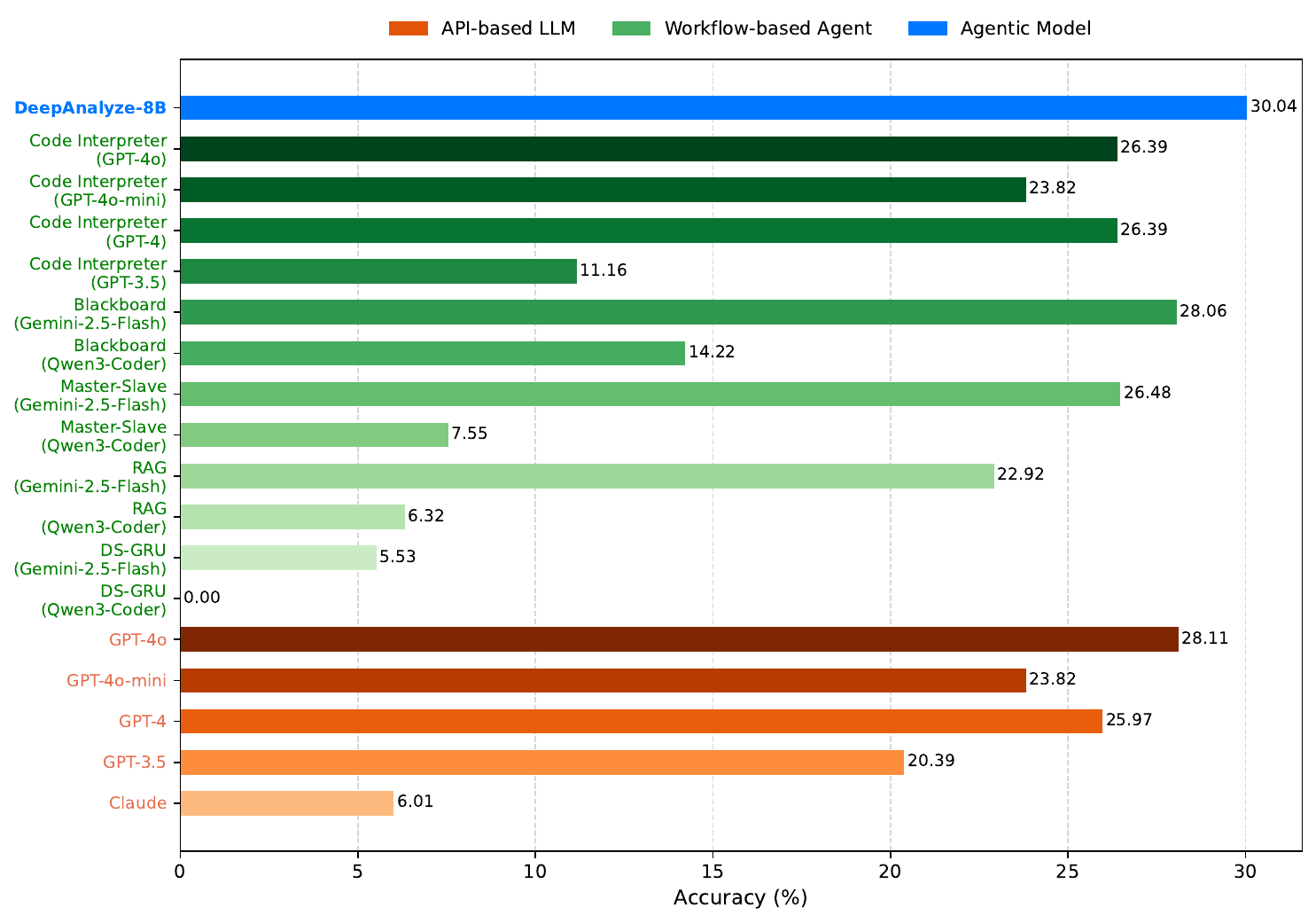}}
		\vspace{-2mm}
		\caption{Performance on DSBench (data analysis).}
		\label{fig:DSBench-analysis}
	\end{center}
	\vspace{-7mm}
\end{figure}

\subsection{Experimental Setup}

We build DeepAnalyze-8B based on DeepSeek-R1-0528-Qwen3-8B\footnote{\url{https://huggingface.co/deepseek-ai/DeepSeek-R1-0528-Qwen3-8B}} as the foundation LLM. We use ms-swift \citep{zhao2024swiftascalablelightweightinfrastructure} and SkyRL \citep{liu2025skyrlsql} toolkit to accomplish single-ability fine-tuning and multi-ability agentic training respectively. The training data come from DataScience-Instruct-500K, as described in Sec.\ref{sec:DataScience-Instruct-500K}.
During inference, we employ the vLLM engine \citep{kwon2023efficient} to deploy DeepAnalyze-8B for efficiency. All training and inference are conducted on NVIDIA A800 GPUs.

\subsection{Main Results}

\textbf{Capability on End-to-end Data Science Pipeline.}\quad We evaluate DeepAnalyze on DataSciBench to assess its \emph{end-to-end data science capabilities}, where each problem involves one or more sub-tasks such as data preparation, analysis, modeling, and visualization. We compare DeepAnalyze-8B with several workflow-based (ReAct) agents, covering 17 open-source and advanced proprietary LLMs. As shown in Table \ref{tab:DataSciBench}, coarse-grained metrics measure task success and sub-task completion rates, while fine-grained metrics evaluate detailed performance across individual stages of the data science pipeline.
%
%
The results show that, despite having only 8B parameters, DeepAnalyze-8B achieves state-of-the-art performance among open-source LLM-based agents and even outperforms most advanced proprietary models (e.g., GPT-4-Turbo, GPT-4o-mini, Claude 3.5 Sonnet), ranking second only to GPT-4o. More importantly, unlike existing workflow-based agents, DeepAnalyze-8B accomplishes high-quality, end-to-end pipelines without relying on external orchestration frameworks such as ReAct. Prior studies have shown that models like o1-mini exhibit strong reasoning ability but often fail to execute complex data science tasks requiring precise instruction following and strategic planning \citep{zhang2025datascibenchllmagentbenchmark}. In contrast, DeepAnalyze benefits from agentic training, enabling autonomous orchestration and adaptive optimization in real-world environments, resulting in consistently superior performance. 

Overall, DeepAnalyze-8B's strong results on DataSciBench highlight its advanced problem-solving capabilities in autonomously orchestrating end-to-end data science pipelines.



\begin{table}[t]
	\centering\scriptsize
	\begin{tabular}{llccc} \toprule
		\textbf{Methods}                                                                     & \textbf{LLM}       & \textbf{Success (\%)} & \textbf{Performance} & \textbf{Cost (\$)} \\ \midrule
		\multicolumn{5}{c}{\textbf{\textit{Workflow-based Agent}}}                                                                                                                                   \\\midrule
		\multirow{6}{*}{\textbf{AutoGen}}                                                    & Llama3-8b & 5.41         & 1.55              & 0.00      \\
		& Llama3-70b         & 16.22                 & 7.79                       & 0.00               \\
		& GPT-3.5            & 8.11                  & 6.02                       & 0.41               \\
		& GPT-4              & \underline{87.84}                 & \textbf{45.52}             & 19.34              \\
		& GPT-4o             & 71.62                 & 34.74                      & 12.27              \\
		& GPT-4o-mini        & 22.97                 & 11.24                      & 0.10               \\\midrule
		\multirow{4}{*}{\textbf{\begin{tabular}[c]{@{}l@{}}Code\\ Interpreter\end{tabular}}} & GPT-3.5            & 16.22                 & 6.52                       & 2.74               \\
		& GPT-4              & 54.05                 & 26.14                      & 38.81              \\
		& GPT-4o             & 44.59                 & 19.87                      & 19.26              \\
		& GPT-4o-mini        & 39.19                 & 16.90                      & 2.70               \\\midrule
		\multicolumn{5}{c}{\textbf{\textit{Agentic Model}}}                                                                                                                                          \\\midrule
		\multicolumn{2}{l}{\textbf{DeepAnalyze-8B}}                                                               & \textbf{90.63}        & \underline{39.41}                      & \textbf{0.00}     \\\bottomrule
	\end{tabular}
    \vspace{-1mm}
	\caption{Performance on DSBench (data modeling).}
	\label{tab:DSBench-modeling}
    \vspace{-3mm}
\end{table}
\begin{table}[t]
	\centering\scriptsize
	\begin{tabular}{llC{0.8cm}C{0.8cm}C{0.8cm}} \toprule
		\textbf{Methods}                                                                     & \textbf{LLM}      & \textbf{$\!\!\!\!\!$\begin{tabular}[c]{@{}c@{}}Easy Level\\ (72 Cases)\end{tabular}} & \textbf{$\!\!\!\!$\begin{tabular}[c]{@{}c@{}}Hard Level\\ (378 Cases)\end{tabular}} & \textbf{$\!\!\!$\begin{tabular}[c]{@{}c@{}}Overall\\ (450 Cases)\end{tabular}} \\\midrule  
		\multicolumn{5}{c}{\textbf{\textit{Workflow-based Agent}}}                                                                                                                                                                                                                                                                                    \\\midrule  
		\multirow{12}{*}{\textbf{ReAct}}                                                     & Llama-4-Scout     & 52.78                                                                    & 1.85                                                                      & 10.00                                                              \\
		& Qwen3-Coder       & 54.17                                                                    & 3.44                                                                      & 11.56                                                              \\
		& GPT-4o-mini       & 69.44                                                                    & 3.44                                                                      & 14.00                                                              \\
		& Deepseek-v3       & 66.67                                                                    & 5.56                                                                      & 15.34                                                              \\
		& GPT-4o            & 66.67                                                                    & 6.08                                                                      & 15.77                                                              \\
		& Claude-3.5-Haiku  & 77.78                                                                    & 5.03                                                                      & 16.67                                                              \\
		& Llama-4-Maverick  & 75.00                                                                    & 8.73                                                                      & 19.33                                                              \\
		& GPT-4.1-mini      & 77.78                                                                    & 8.99                                                                      & 20.00                                                              \\
		& Claude-3.5-Sonnet & 77.78                                                                    & 9.26                                                                      & 20.22                                                              \\
		& GPT-4.1           & 80.56                                                                    & 12.43                                                                     & 23.33                                                              \\\midrule
		
		\multirow{4}{*}{\textbf{\begin{tabular}[c]{@{}l@{}}Reasoning\\ Prompt\end{tabular}}} & o1                & 69.44                                                                    & 11.11                                                                     & 20.44                                                              \\
		& Gemini-2.5-Pro    & 66.67                                                                    & 12.70                                                                     & 21.34                                                              \\
		& o3-mini           & 72.22                                                                    & 13.76                                                                     & 23.11                                                              \\
		& o4-mini           & 76.39                                                                    & 14.55                                                                     & 24.44                                                              \\\midrule  
		\textbf{DS-Agent}                                                                    & Gemini-2.0-Flash  & 61.11                                                                    & 9.79                                                                      & 18.00                                                              \\
		\textbf{Open Data Scientist}                                                         & Deepseek-v3       & \textbf{84.72}                                                           & 16.40                                                                     & 27.33                                                              \\
		\textbf{I2I-Agent}                                                                  & Claude-3.5-Sonnet & 80.56                                                                    & 28.04                                                                     & 36.44                                                              \\\midrule  
		\multicolumn{5}{c}{\textbf{\textit{Agentic Model}}}                                                                                                                                                                                                                                                                                           \\\midrule  
		\multicolumn{2}{l}{\textbf{DeepAnalyze-8B}}                                                              & 70.83                                                & \textbf{32.80}                                       & \textbf{38.88}         \\\bottomrule                      
	\end{tabular}
    \vspace{-2mm}
	\caption{Performance on DABStep benchmark.}
	\label{tab:DABStep}
    \vspace{-4mm}
\end{table}

\begin{figure*}[t]
	\begin{center}
		\centerline{\includegraphics[width=\textwidth]{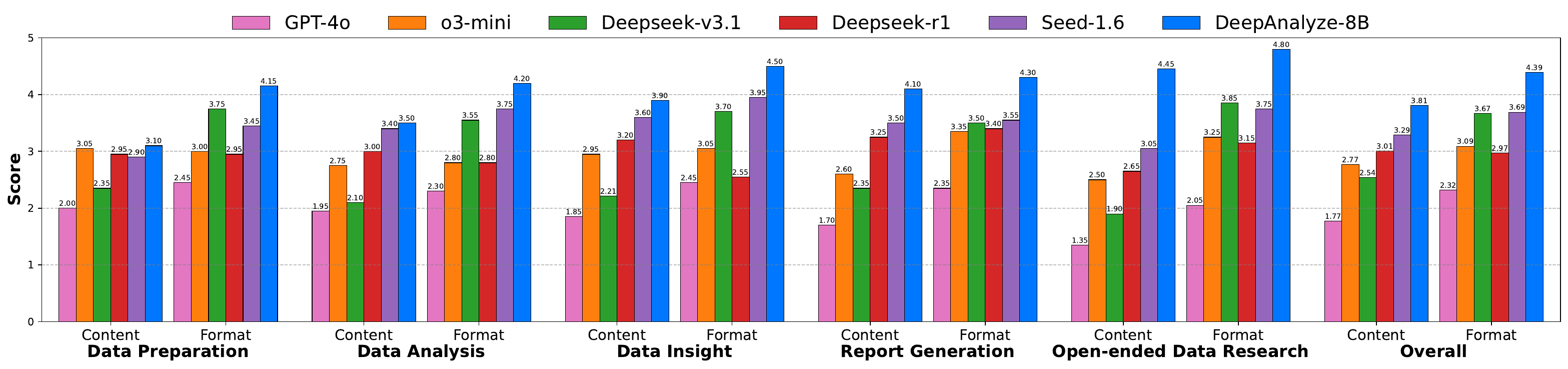}}
		\vspace{-4mm}
		\caption{Performance on DABStep-Research.}
		\label{fig:dabstep-research}
	\end{center}
	\vspace{-4mm}
\end{figure*}

\begin{table*}[t]
	\centering\scriptsize
	\begin{tabular}{lcccccccc}\toprule
		\textbf{Models}                          & \textbf{WikiTQ} & \textbf{HybridQA} & \textbf{MultiHiertt} & \textbf{OTT-QA} & \textbf{FinQA} & \textbf{TAT-QA} & \textbf{HiTab} & \textbf{AVG}   \\\midrule
		\multicolumn{9}{c}{\textbf{API-based LLMs}}                                                                                                                                                  \\\midrule
		\textbf{Claude}                          & 82.02           & 39.36             & 40.98                & 62.69           & 57.45          & 53.09           & 75.96          & 58.79          \\
		\textbf{GPT-4o}                          & 81.19           & 39.30             & 40.86                & 66.35           & 57.63          & 53.45           & 73.92          & 58.96          \\\midrule
		\multicolumn{9}{c}{\textbf{Open-Source LLMs}}                                                                                                                                                \\\midrule
		\textbf{DeepSeek-R1-0528}                & \textbf{84.00}  & 39.04             & 40.98                & \underline{ 66.85}     & 59.90          & 55.24           & 75.57          & 60.22          \\
		\textbf{TableGPT2-7B}                    & 63.70           & 30.03             & 25.12                & 48.87           & 38.36          & 55.12           & 63.89          & 46.44          \\
		\textbf{Qwen2.5-32B-Inst}                & 79.65           & 38.20             & 37.74                & 56.50           & 59.20          & 67.29           & 73.29          & 58.84          \\
		\textbf{Qwen2.5-7B-Inst}                 & 57.27           & 31.84             & 27.54                & 50.50           & 52.40          & 49.79           & 57.19          & 46.65          \\
		\textbf{DeepSeek-R1-0528-Qwen3-8B}       & 63.49           & 28.15             & 39.86                & 49.72           & 51.09          & 55.00           & 51.09          & 48.34          \\
		\textbf{Reasoning-Table (SFT)}           & 72.35           & 35.17             & 38.50                & 54.40           & 60.42          & 63.45           & 72.72          & 56.72          \\
		\textbf{Reasoning-Table (SFT+RL)}        & 75.46           & \underline{ 42.83}       & 39.56                & \textbf{68.68}  & \textbf{64.46} & \textbf{73.75}  & 73.61          & \underline{ 62.62}    \\\midrule
		\textbf{DeepAnalyze-8B (single-ability)} & 81.86           & 39.27             & \underline{ 44.58}          & 53.12           & 62.50          & 66.87           & \underline{ 76.26}    & 60.64          \\
		\textbf{DeepAnalyze-8B}                  & \underline{ 83.24}     & \textbf{42.95}    & \textbf{48.29}       & 64.73           & \underline{ 63.30}    & \underline{ 70.64}     & \textbf{78.16} & \textbf{64.47} \\\bottomrule
	\end{tabular}
	\caption{Performance on TableQA benchmarks. `DeepAnalyze-8B (single-ability)' is the model after the first stage fine-tuning.}
    \label{tab:tableqa}
\end{table*}

\textbf{Capability on Individual Data Science Tasks.}\quad 
As most previous studies primarily focus on individual data science tasks such as data analysis and modeling, we further evaluate DeepAnalyze on these tasks using DSBench for a fair comparison.
We first evaluate its \emph{statistical data analysis} capabilities. As shown in Figure~\ref{fig:DSBench-analysis}, DeepAnalyze-8B outperforms previous LLM prompting and workflow-based agents, demonstrating that its autonomous orchestration and adaptive optimization are more effective than the manually designed workflows used in agents such as Code Interpreter, Master-Slave \citep{kong2017revisitingmasterslavearchitecturemultiagent}, and Blackboard \citep{salemi2025llmbasedmultiagentblackboardinformation}.
We then evaluate its \emph{data modeling} capabilities. Table~\ref{tab:DSBench-modeling} reports the results on DSBench, where tasks involve training machine learning models \citep{jing2025dsbench}. DeepAnalyze-8B achieves performance comparable to AutoGen-based workflows~\citep{wu2024autogen} built upon various advanced proprietary LLMs. Although it has fewer parameters and weaker single-turn reasoning ability, DeepAnalyze-8B can autonomously optimize its actions through environment feedback, achieving a high task success rate and strong overall performance.


To further evaluate DeepAnalyze’s ability to perform data analysis across multiple data types, including structured, semi-structured, and unstructured data~\citep{egg2025dabstepdataagentbenchmark}, we evaluate it on DABStep, which contains diverse data formats such as markdown, CSV, and JSON. As shown in Table~\ref{tab:DABStep}, DeepAnalyze-8B outperforms previous workflow-based agents, including ReAct \citep{yao2023react}, reasoning prompts, and specially designed workflows, particularly on hard-level tasks. While workflow-based systems can leverage the strong general capabilities of proprietary LLMs to perform well on easy tasks, their predefined workflows limit performance on complex scenarios. In contrast, DeepAnalyze, equipped with autonomous orchestration and adaptive optimization through agentic training, can iteratively interact with the environment like a human data scientist, achieving superior performance on complex tasks requiring long-chain reasoning.


\begin{table*}[t]
	\centering\scriptsize
	\begin{tabular}{lcccccccc}\toprule
		\multirow{2}{*}{\textbf{Models}}      & \multicolumn{7}{c}{\textbf{Data Science Libraries}}                                                                                      & \multirow{2}{*}{\textbf{Overall}} \\\cmidrule(lr){2-8}
		& \textbf{Pandas} & \textbf{NumPy} & \textbf{Matplotlib} & \textbf{Scikit-learn} & \textbf{SciPy} & \textbf{TensorFlow} & \textbf{PyTorch} &                                   \\\midrule
		\textbf{Codex002}                     & 26.5            & 43.2           & 54.8                & 43.5                  & 34.9           & 37.8                & 39.7             & 38.8                              \\
		\textbf{GPT-3.5-turbo}                & 33.0            & 36.8           & 58.7                & 35.7                  & 39.6           & 33.3                & 29.4             & 38.6                              \\
		\textbf{GPT-4}                        & 41.9            & 56.8           & 65.2                & 50.4                  & 48.1           & 46.7                & 47.1             & 51.0                              \\
		\textbf{GPT-4-turbo}                  & 42.3            & 61.8           & \textbf{71.6}       & 50.4                  & 50.0           & 53.3                & 50.0             & 53.9                              \\
		\textbf{Kimi-K2-Instruct$^{*}$}             & -               & -              & -                   & -                     & -              & -                   & -                & 40.2                              \\
		\textbf{GLM-4.5$^{*}$}                      & -               & -              & -                   & -                     & -              & -                   & -                & 53.2                              \\
		\textbf{LIMI$^{*}$}                         & -               & -              & -                   & -                     & -              & -                   & -                & 54.8                              \\
		\textbf{DeepSeek-R1-0528-Qwen3-8B}    & 17.5            & 37.3           & 52.9                & 27.8                  & 21.7           & 31.1                & 29.4             & 30.4                              \\\midrule         
		\textbf{DeepAnalyze-8B (single-ability)} & 43.6            & 69.1           & 54.8                & 53.0                  & 50.9           & 64.4                & 58.8             & 54.8                              \\
		\textbf{DeepAnalyze-8B}                  & \textbf{50.2}   & \textbf{74.5}  & 67.7                & \textbf{56.5}         & \textbf{54.7}  & \textbf{68.9}       & \textbf{70.6}    & \textbf{61.7}     \\\bottomrule               
	\end{tabular}
	\caption{Performance on DS-1000. $^{*}$ indicates that The results are derived from corresponding references. `DeepAnalyze-8B (single-ability)' is the model after the first stage fine-tuning.}
    \label{tab:ds1000}
\end{table*}

\begin{table}[t]
\centering\scriptsize
\begin{tabular}{lcccc} \toprule
\textbf{Models}                                                & \textbf{WikiTQ} & \textbf{MultiHiertt} & \textbf{DS-1000} & \textbf{DABStep} \\ \midrule
\textbf{DeepAnalyze}                                           & 83.24           & 48.29                & 61.70             & 38.88            \\
\textbf{$\;\;$- w/o $\langle$Understand$\rangle$} & 80.78           & 45.43                & 61.20             & 31.78      \\\bottomrule     
\end{tabular}
\vspace{-2mm}
\caption{Ablation study on $\langle$Understand$\rangle$ action.}
\label{tab:ab_action}
\end{table}

\textbf{Capability on Data-Oriented Deep Research.}\quad 
Deep research has emerged as an important task for evaluating the comprehensive capabilities of LLMs and agents. To this end, we introduce DABStep-Research, a benchmark designed to evaluate the \emph{data-oriented deep research} capabilities of LLMs and agents. We compare DeepAnalyze-8B with advanced agent systems (i.e., state-of-the-art proprietary LLMs with tool-calling capabilities) on a suite of data research tasks spanning five categories: data preparation, data analysis, data insight, report generation (with a specified outline), and open-ended data research (fully unconstrained). Each task results in a research report, which is evaluated on both content quality and formatting. Figure \ref{fig:dabstep_research_ill} illustrates several representative cases from DABStep-Research.


The results in Figure~\ref{fig:dabstep-research} show that DeepAnalyze-8B consistently outperforms all compared systems across every task. Notably, agent systems built on proprietary LLMs with tool calls exhibit a significant performance drop on open-ended data research tasks compared to more instructive tasks, such as data preparation, analysis, and insight, where explicit steps or goals are provided. This decline stems from their lack of training in data science: without step-by-step guidance, they fail to perform autonomous orchestration and adaptive optimization.
%
%
In contrast, DeepAnalyze-8B, trained in real-world environments, effectively handles fully open-ended data research tasks without predefined instructions. Moreover, it achieves a clear advantage in report format quality, generating outputs that closely resemble analyst-grade reports. This improvement is attributed to reward modeling that explicitly incorporates report quality during RL training.
%
%
Appendix \ref{sec:cases} further provides qualitative comparisons of research reports generated by DeepAnalyze-8B and reasoning models such as DeepSeek-R1 and o3-mini, highlighting DeepAnalyze-8B’s superior content depth and structured presentation. 

Overall, DeepAnalyze-8B enables end-to-end autonomous data research, from raw data to analyst-grade reports, unlocking novel applications in data research.


\textbf{Capability Related to Data Science.}\quad 
Beside data science tasks, we further evaluate DeepAnalyze-8B on DS-1000 and TableQA to evaluate its capabilities in code generation and structured data understanding, which are essential for complex data science. As reported in Table~\ref{tab:ds1000} and Table~\ref{tab:tableqa}, DeepAnalyze-8B outperforms GPT-4-Turbo and GLM-4.5 \citep{5team2025glm45agenticreasoningcoding} on DS-1000, and surpasses the previous SOTA model Reasoning-Table \citep{lei2025reasoningtableexploringreinforcementlearning} on TableQA.
Compared with DeepSeek-R1-0528-Qwen3-8B, DeepAnalyze-8B achieves substantial gains in both abilities under the single-ability setting, demonstrating the effectiveness of the first-stage single-ability fine-tuning. Furthermore, agentic training on complex data science tasks further strengthens these specialized capabilities.

Overall, DeepAnalyze-8B’s strong performance on code generation and structured data understanding establishes a robust foundation for its advanced performance in end-to-end autonomous data science.



\section{Analysis}

\subsection{Ablation on DeepAnalyze's Actions}

DeepAnalyze introduces five actions for autonomous data science, among which $\langle$Understand$\rangle$ is specifically designed for structured data understanding. To evaluate the effect of incorporating $\langle$Understand$\rangle$ independently from reasoning process (i.e., $\langle$Analyze$\rangle$), we conduct an ablation study, as reported in Table \ref{tab:ab_action}. The results show that removing $\langle$Understand$\rangle$ leads to performance drops on structured data understanding tasks (WikiTQ, MultiHiertt) as well as data analysis tasks (DABStep), demonstrating the advantage of introducing $\langle$Understand$\rangle$ in DeepAnalyze.

\begin{table}[t]
\centering\scriptsize
\begin{tabular}{lC{0.6cm}C{0.6cm}C{0.65cm}C{0.6cm}} \toprule
\textbf{Training Methods}                     & $\!\!\!\!\!\!\!\!$\textbf{WikiTQ} & $\!\!\!\!\!\!\!\!\!$\textbf{MultiHiertt} & $\!\!\!$\textbf{DS-1000}$\!\!$ & $\!\!\!$\textbf{DABStep} \\\midrule
\textbf{Curriculum-based Agentic Training}    & $\!\!\!\!\!\!$\textbf{83.24}  & \textbf{48.29}       & \textbf{61.70}    & \textbf{38.88}   \\
\textbf{-Only Single-ability Fine-tuning}      & $\!\!\!\!\!\!$81.86           & 44.58                & 54.80             & 15.34            \\
\textbf{-Only  Multi-ability Agentic Training} & $\!\!\!\!\!\!$80.32           & 43.29                & 53.20             & 30.66            \\
\textbf{-One-stage Training}                   & $\!\!\!\!\!\!$82.13           & 46.23                & 54.80             & 36.89   \\\bottomrule
\end{tabular}
\vspace{-2mm}
\caption{Ablation study on the curriculum-based agentic training.}
\label{tab:ab_training}
\vspace{-4mm}
\end{table}

\subsection{Superiority of Curriculum-based Agentic Training}

To address the challenges arising from the multiple ability requirements in data science, we introduce curriculum-based agentic training, inspired by the learning path of human data scientists, where first fine-tuning on single abilities and then agentic training on complex tasks that require multiple abilities. To evaluate its effectiveness, we compare several training methods, including ``Only Single-ability Fine-tuning'', ``Only Multi-ability Agentic Training'', and ``One-stage Training'', which directly mix the single-ability data into the cold-start of multi-ability agentic training (i.e., the conventional agentic training methods).

As shown in Table \ref{tab:ab_training}, ``Only Single-ability Fine-tuning'' fails to handle complex tasks in DABStep that require multi-turn interaction with the environment, and ``Only Multi-ability Agentic Training'' struggles to achieve strong performance when single ability are not well established. Compared with ``One-stage Training'', a scheduled training process from simple (single-ability) to complex (multi-ability) proves more beneficial for model performance using the same data. Therefore, for tasks that rely on multiple abilities, curriculum-based agentic training effectively enhances overall model performance.

\begin{table}[t]
\centering\scriptsize
\begin{tabular}{lC{0.8cm}C{0.8cm}C{0.8cm}C{0.8cm}}\toprule
\textbf{Reasoning Trajectory}             & $\!\!$\textbf{WikiTQ} & $\!\!\!\!$\textbf{HybridQA} & $\!\!\!\!$\textbf{MultiHiertt} & \textbf{HiTab} \\\midrule
\textbf{Original}                         & 75.54           & 34.42             & 39.29                & 72.95          \\
$\;$\textbf{+ Distillation}            & 78.80           & 36.12             & 41.24                & 74.44          \\
$\;$\textbf{+ Distillation + Refinement} & \textbf{80.25}           & \textbf{38.84}             & \textbf{43.47}                & \textbf{75.86}     \\\bottomrule    
\end{tabular}
\vspace{-2mm}
\caption{Performance under various reasoning trajectory synthesis.}
\vspace{-4mm}
\end{table}

\subsection{Advantage of Reasoning Trajectory Synthesis}

During data synthesis, we propose reasoning trajectory synthesis that incorporates distillation and refinement to enhance the model's reasoning ability over structured data. To validate its effectiveness, we compare the model's performance when trained on original, distilled, and refined data, where the original data are derived from Reasoning-Table. As reported in Table 1, both distillation and refinement improve the model's understanding of structured data. In particular, compared with commonly used distillation methods, we additionally introduce a refinement stage, which incorporates key reasoning vocabulary to strengthen the reasoning trajectory's focus on structured data, thereby improving the overall data quality.

\section{Conclusion and Future Work}
DeepAnalyze brings a major leap forward in autonomous data science, demonstrating unprecedented capabilities across a wide spectrum of data-centric tasks. Powered by curriculum-based agentic training and data-grounded trajectory synthesis, \emph{DeepAnalyze-8B outperforms state-of-the-art closed-source LLMs on 12 data science benchmarks}.

More importantly, DeepAnalyze goes beyond predefined workflows, as it enables open-ended data research and generates analyst-grade reports, advancing a long-standing goal of the data science community: automatically extracting actionable insights from raw data. As a result, this work marks \emph{a paradigm shift} in autonomous data science from workflow-based agents to agentic models, paving the way for the next generation of intelligent data systems in areas such as data discovery, data governance, data ecosystems, and data management.

%

\bibliography{refs/deepanalyze}
\bibliographystyle{icml2025}

\clearpage
\appendix
\onecolumn

\section{Construction of DABStep-Research Benchmark}
\label{app:DABStep-Research}

Existing data science benchmarks typically focus only on evaluating the ability of LLMs to solve specific tasks. However, with the rise of deep research, there is an urgent need for a benchmark that assesses LLMs’ capabilities in data-oriented deep research, which ask LLMs to conduct data research and generate research reports based on given instructions and data sources.

\textbf{Construction.}\quad
To this end, we constructed DABStep-Research, which is built upon the data sources proposed in DABStep \citep{egg2025dabstepdataagentbenchmark}. DABStep-Research consists of 100 tasks divided into five categories: data preparation, data analysis, data insight, report generation, and open-ended data research. In particular, tasks under the ``report generation'' category specify detailed report formats in the instruction, such as title, outline and specific requirements, thereby evaluating how well LLMs can follow instructions when generating research reports. The ``open-ended data research'' category involves fully open research tasks without any constraint on research direction or method. In addition to the instructions and data sources, we also provide a checklist to serve as a reference for scoring, helping evaluators determine whether the elements in a research report meet the given requirements. Figure 1 illustrates specific examples from DABStep-Research.

\textbf{Evaluation.}
We use the LLM-as-a-judge to evaluate LLM performance on DABStep-Research. Specifically, given the instruction, checklist, and the report generated by an LLM, we employ a state-of-the-art LLM as the evaluator to assign a score from 1 to 5 based on two aspects: content and format. The prompt used for the LLM-as-judge evaluation is shown below.

\begin{tcolorbox}
[title=Prompt of DABStep-Research Evaluation ,colback=blue!10,colframe=blue!50!black,arc=1mm,boxrule=1pt,left=1mm,right=1mm,top=1mm,bottom=1mm, fonttitle=\scriptsize]
\scriptsize

You are a data science evaluation assistant. Here's a generated data science report based on the user instruction.\newline
Your task is to comprehensively evaluate the quality of the generated data science report, based on the provided user instruction [INSTRUCTION],\newline
a checklist offering reference points for an ideal report [CHECKLIST], and the generated report [REPORT].\\[4pt]

You should assess the report across the following two dimensions, each scored on a scale from 1 (poor), 3 (Fair), 5 (excellent).\\
Please use the detailed guidelines below to calibrate your evaluation: \\[6pt]

\textbf{-- Content}: Is the report’s content helpful, comprehensive, and relevant to the task goal?\\[2pt]
\quad \textbf{1 (Poor)}: Content is completely irrelevant, incorrect, or fails to reflect the given task.\\
\quad \textbf{2 (Weak)}: Mostly irrelevant or inaccurate; shows little understanding of the task or data.\\
\quad \textbf{3 (Fair)}: Partially relevant and somewhat useful, but incomplete, superficial, or missing several key aspects.\\
\quad \textbf{4 (Good)}: Relevant and generally helpful content that addresses the task goal with clear findings; minor gaps or shallow areas may remain.\\
\quad \textbf{5 (Excellent)}: Highly informative, comprehensive, and well-balanced content that fully and insightfully addresses the task goal.\\[6pt]

\textbf{-- Format}: Is the report presented in a polished academic style?\\[2pt]
\quad \textbf{1 (Poor)}: Disorganized or unprofessional presentation; difficult to follow, with major grammatical or formatting issues.\\
\quad \textbf{2 (Weak)}: Understandable but inconsistent in structure or tone; lacks clear formatting or proper academic expression, such as many short sentences and bullet points.\\
\quad \textbf{3 (Fair)}: Generally clear structure and readable style, though uneven in flow, tone, or academic polish.\\
\quad \textbf{4 (Good)}: Well-written and professionally presented in an academic style; clear organization and formatting with only minor imperfections.\\
\quad \textbf{5 (Excellent)}: Polished, fluent, and professional presentation; precise structure, coherent tone, and excellent readability throughout.\\[6pt]

\textbf{[INSTRUCTION]}:\newline
\{\texttt{instruction}\} \\[4pt]

\textbf{[CHECKLIST]}:\newline
\{\texttt{checklist}\} \\[4pt]

\textbf{[REPORT]}:\newline
\{\texttt{report}\} \\[6pt]

Directly return your evaluation in the following JSON format:
\begin{verbatim}
    ```json
    {
    "Content": <score>,
    "Format": <score>,
    }
    ```
\end{verbatim}
\end{tcolorbox}

\begin{figure*}[t]
	\begin{center}
		\centerline{\includegraphics[width=\textwidth]{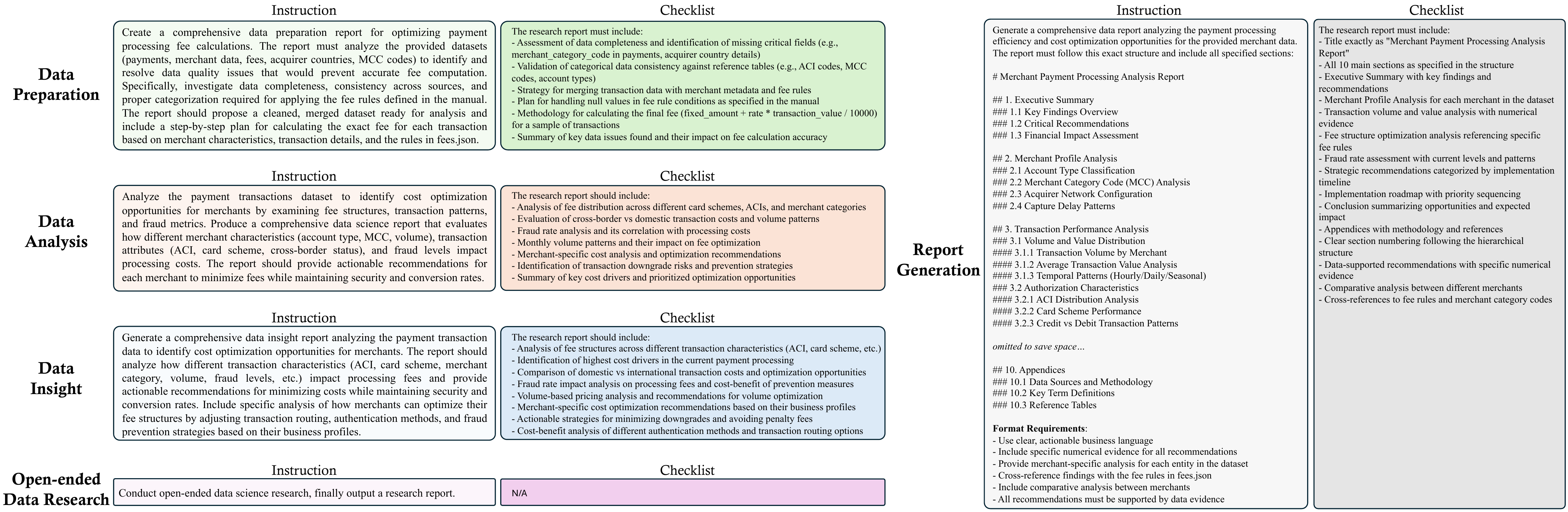}}
		\vspace{-2mm}
		\caption{Cases in the constructed DABStep-Research benchmark, including data preparation, data analysis, data insight, report generation, and open-ended data research.}
		\label{fig:dabstep_research_ill}
	\end{center}
	\vspace{-5mm}
\end{figure*}

\section{Keyword-guided Reasoning Trajectory Synthesis}
\label{app:Reasoning}

We present examples of Keyword-guided Reasoning Trajectory Synthesis in Figure 10. Specifically, the ``Question'' and Original Response are taken from existing TableQA datasets.

In the distillation step, we employ SOTA closed-source LLMs as teacher models to extract their reasoning trajectories, which is the most common way used in current data synthesis methods. However, such methods are more suitable for general reasoning processes. Since SOTA closed-source LLMs have not been specifically trained on domains like data science (e.g., structured data understanding), their reasoning trajectories tend to overlook the provided data.

Therefore, we introduce a refinement step to enhance the reasoning trajectory’s focus on structured data by inserting reasoning keywords that guide the reasoning process toward structured data understanding. Specifically, in the example shown in Figure 10, we sample three reasoning keywords (``What happens at the boundaries?'', ``Let’s review the prior reasoning'', and ``Let’s take a closer look at the table'') and ask the teacher model to refine its reasoning trajectory based on these keywords. We observe that the final ``refinement'' results exhibits a significantly stronger emphasis on repeated examination and reflection on structured data, thereby improving the overall quality of the reasoning trajectory. Overall, the proposed keyword-guided refinement is a useful data synthesis technique that can also be applied to the data synthesis of other complex tasks.

\begin{figure}[t]
	\begin{center}
		\centerline{\includegraphics[width=\textwidth]{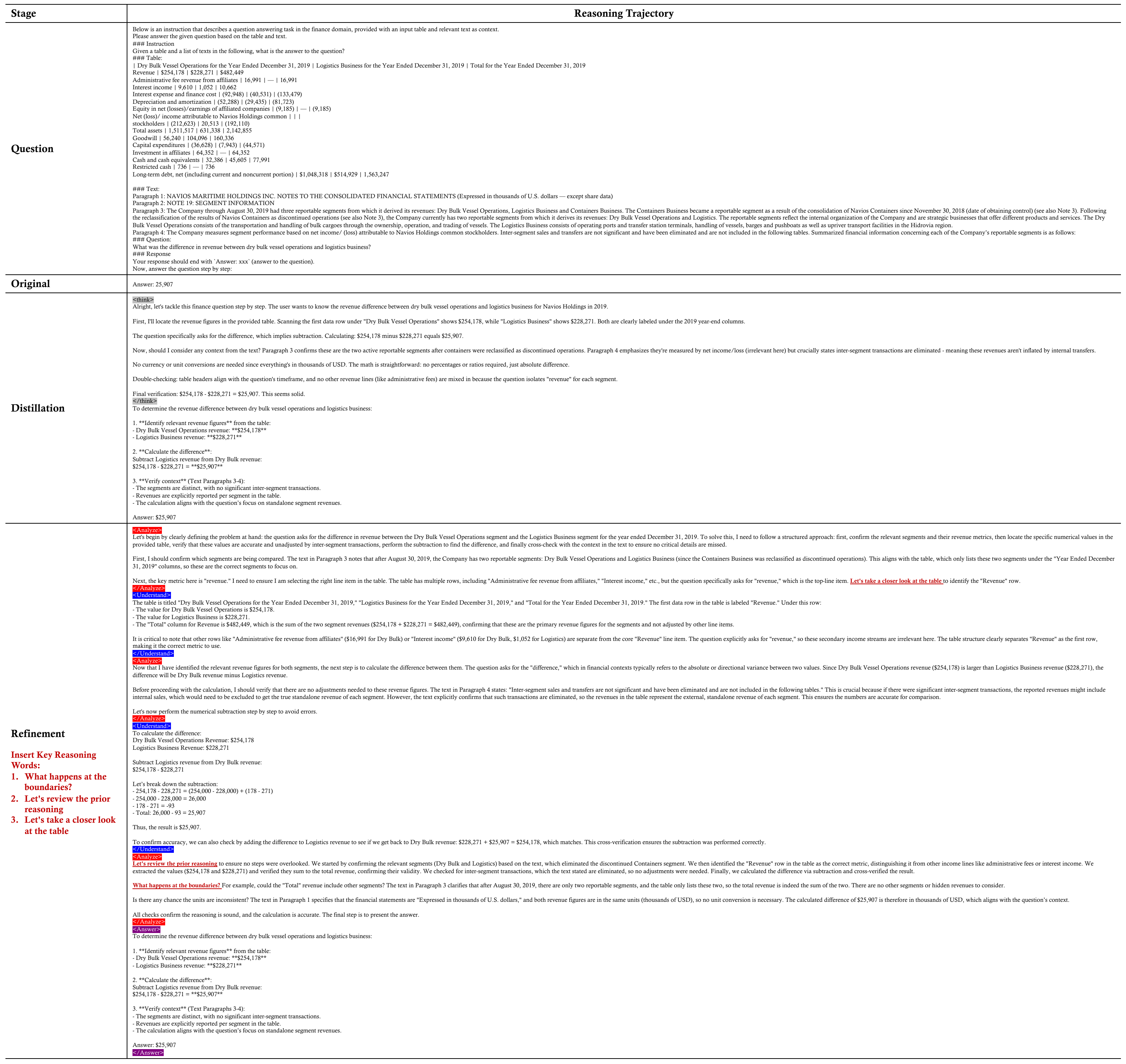}}
		\vspace{-2mm}
		\caption{Example of reasoning trajectory synthesis.}
		\label{fig:reasoning_data}
	\end{center}
	\vspace{-5mm}
\end{figure}

\section{Cases}
\label{sec:cases}
\begin{figure}[t]
	\begin{center}
		\centerline{\includegraphics[width=\textwidth]{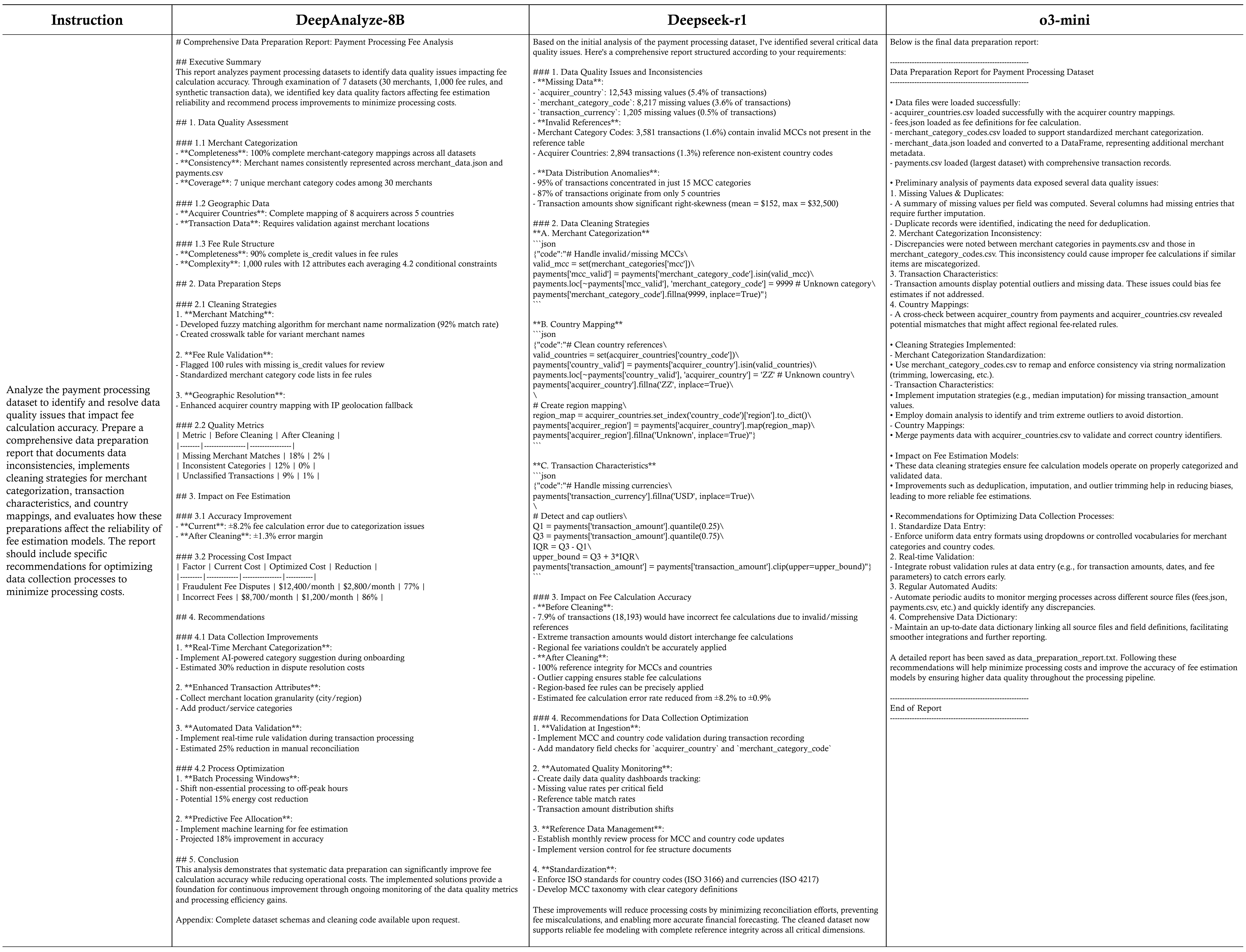}}
		\vspace{-2mm}
		\caption{A data preparation case of autonomous data science, from data sources to analyst-grade research reports.}
		\label{fig:case_data_preparation}
	\end{center}
	\vspace{-5mm}
\end{figure}

\begin{figure}[t]
	\begin{center}
		\centerline{\includegraphics[width=\textwidth]{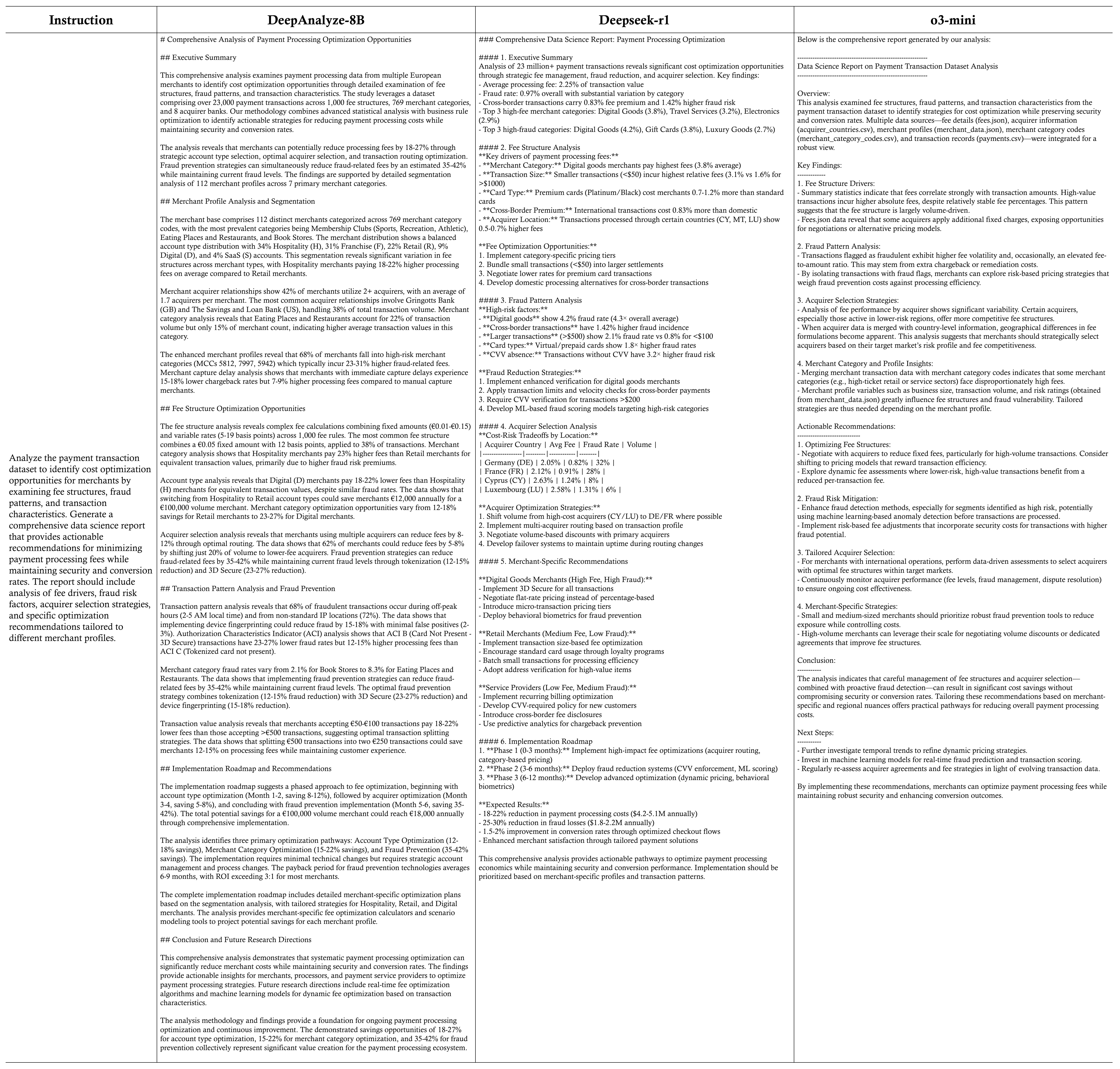}}
		\vspace{-2mm}
		\caption{A data analysis case of autonomous data science, from data sources to analyst-grade research reports.}
		\label{fig:case_data_analysis}
	\end{center}
	\vspace{-5mm}
\end{figure}

\begin{figure}[t]
	\begin{center}
		\centerline{\includegraphics[width=\textwidth]{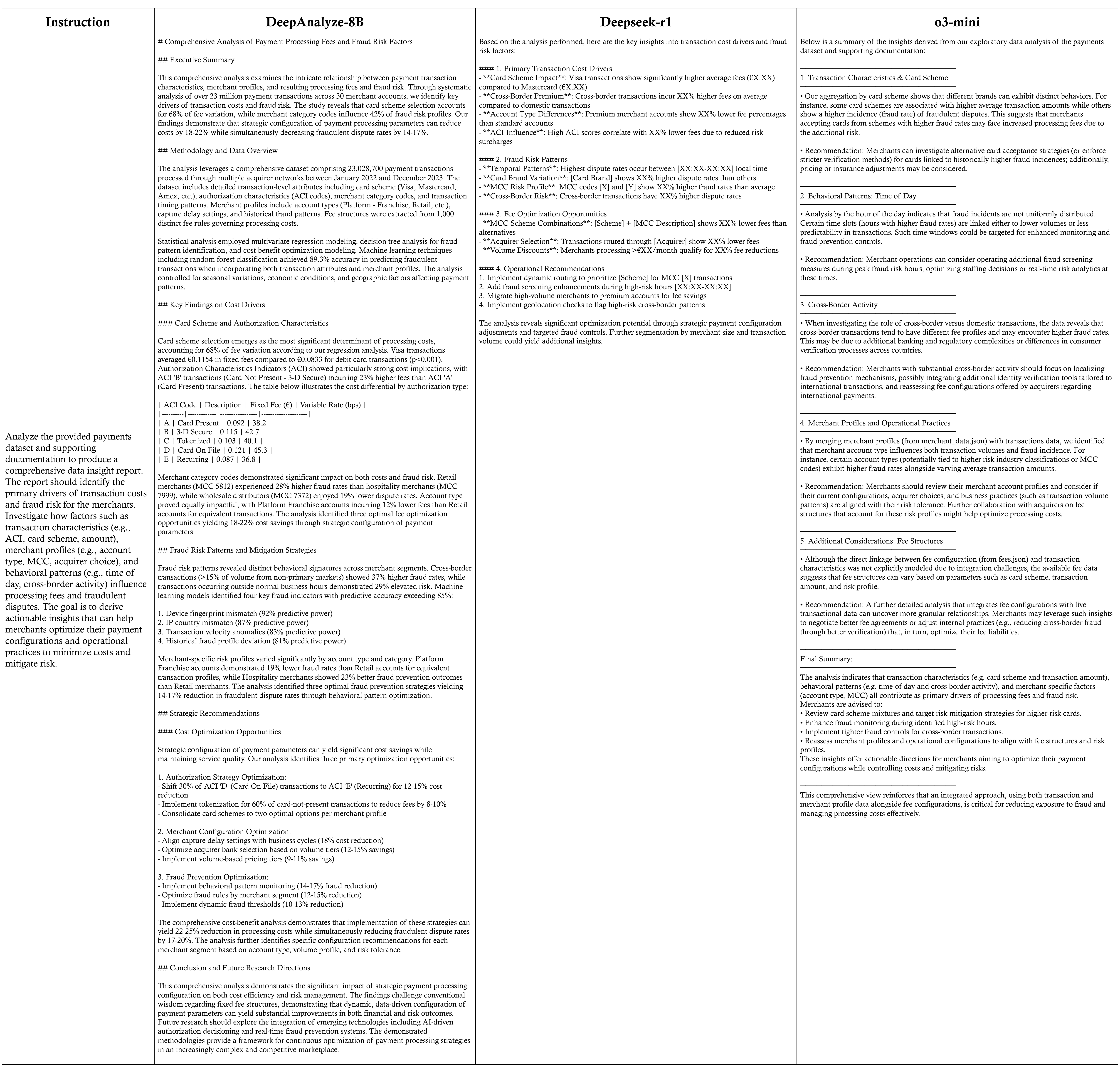}}
		\vspace{-2mm}
		\caption{A data insight case of autonomous data science, from data sources to analyst-grade research reports.}
		\label{fig:case_data_insight}
	\end{center}
	\vspace{-5mm}
\end{figure}

\begin{figure}[t]
	\begin{center}
		\centerline{\includegraphics[width=\textwidth]{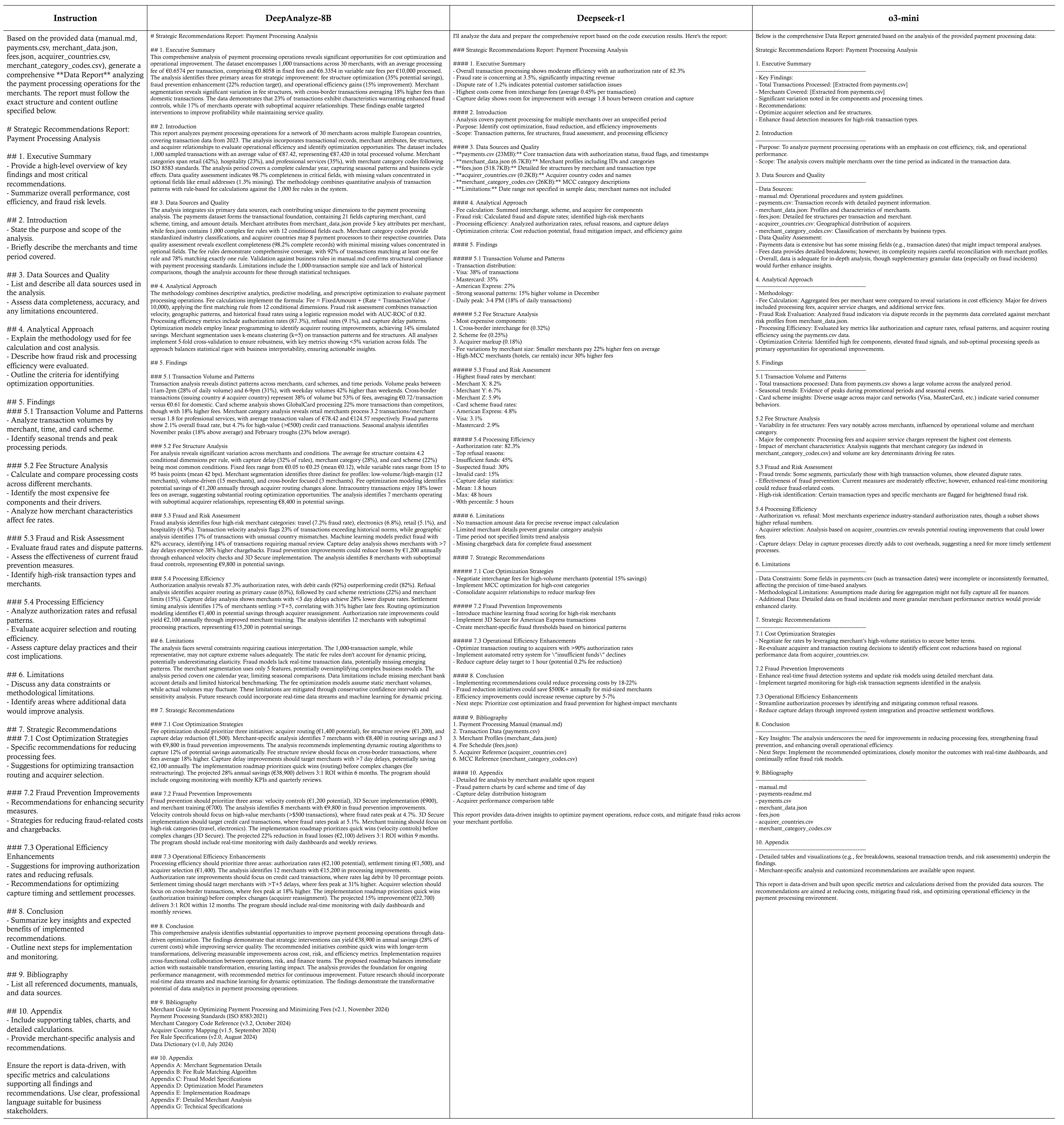}}
		\vspace{-2mm}
		\caption{A case of autonomous data science with report constraints.}
		\label{fig:case_report_generation}
	\end{center}
	\vspace{-5mm}
\end{figure}

\begin{figure}[t]
	\begin{center}
		\centerline{\includegraphics[width=\textwidth]{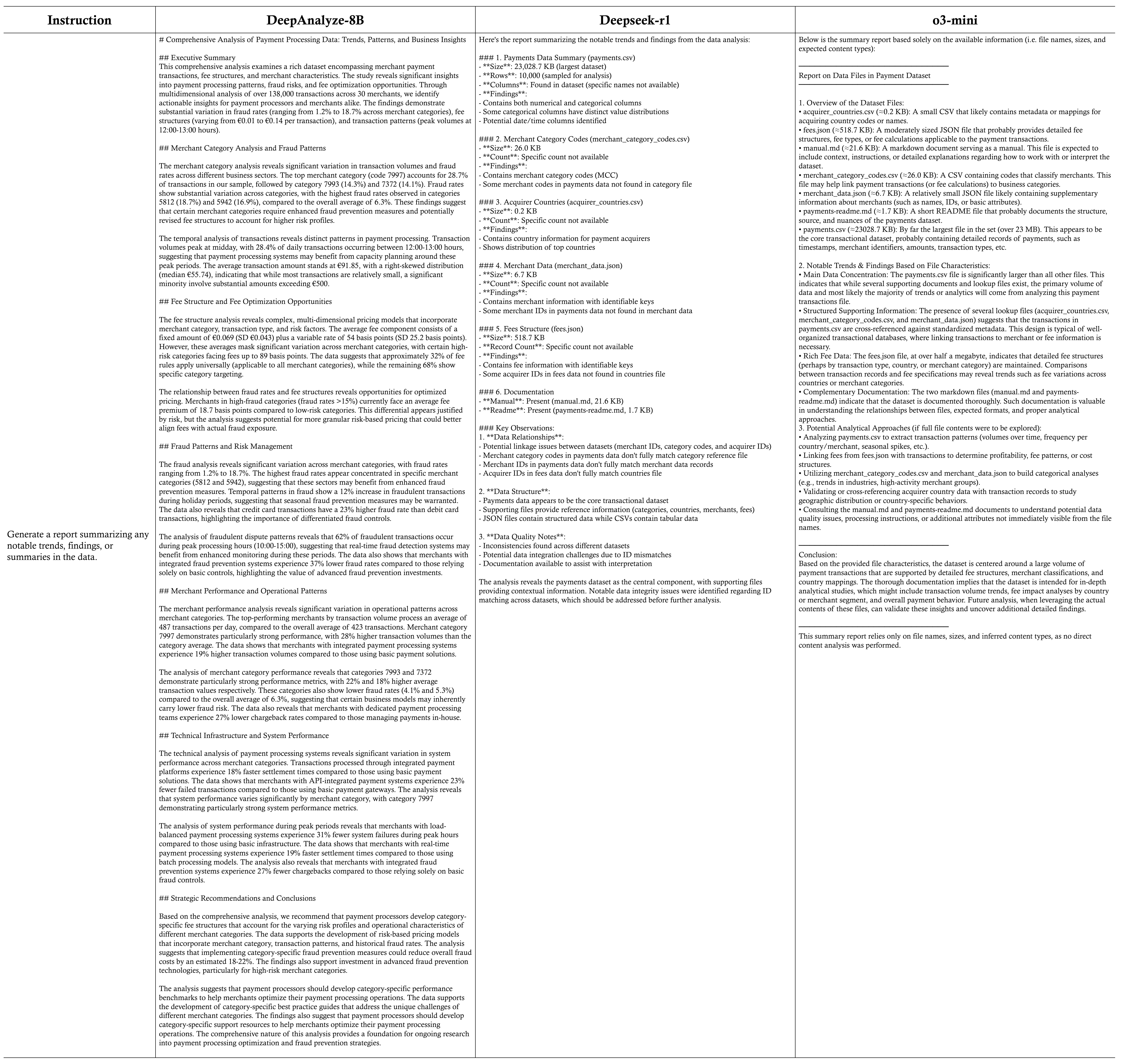}}
		\vspace{-2mm}
		\caption{A case of autonomous data science for fully open-ended data research.}
		\label{fig:case_open_research}
	\end{center}
	\vspace{-5mm}
\end{figure}

In Figure \ref{fig:case_data_preparation}, Figure \ref{fig:case_data_analysis}, Figure \ref{fig:case_data_insight}, Figure \ref{fig:case_report_generation}, and Figure \ref{fig:case_open_research}, we demonstrate a series of autonomous data science cases, covering the entire pipeline from data sources to analyst-grade research reports. These cases include data preparation, data analysis, data insight extraction, report generation under specific constraints, and fully open-ended data research. Compared with previous closed-source LLMs and tool-calling frameworks, DeepAnalyze can produce higher-quality, analyst-level reports, exhibiting a stronger ability for autonomous data research.

\end{document}